\documentclass[9.5pt,cspaper,compsoc,final]{IEEEtran}
\usepackage{customized}

\usepackage{ragged2e}
\usepackage{epsfig}
\usepackage{graphicx}
\usepackage{amsmath}
\usepackage{amssymb}
\usepackage{booktabs}
\usepackage{tabulary}
\usepackage[dvipsnames]{xcolor}
\usepackage[caption=false]{subfig}
\usepackage{soul}
\usepackage{xspace}\xspace
\usepackage{adjustbox}
\usepackage{array}
\usepackage{dblfloatfix}
\usepackage{transparent}
\usepackage{algorithm}
\usepackage{listings}

\newcommand{\revision}[1]{\textcolor{black}{#1}}
\newcommand{\blank}[1]{{#1}}

\usepackage{paralist}
\let\itemize\compactitem
\let\enditemize\endcompactitem
\let\enumerate\compactenum
\let\endenumerate\endcompactenum

\usepackage{etoolbox}
\makeatletter
\AfterEndEnvironment{algorithm}{\let\@algcomment\relax}
\AtEndEnvironment{algorithm}{\kern2pt\hrule\relax\vskip3pt\@algcomment}
\let\@algcomment\relax
\newcommand\algcomment[1]{\def\@algcomment{\footnotesize#1}}
\renewcommand\fs@ruled{\def\@fs@cfont{\bfseries}\let\@fs@capt\floatc@ruled
	\def\@fs@pre{\hrule height.8pt depth0pt \kern2pt}%
	\def\@fs@post{}%
	\def\@fs@mid{\kern2pt\hrule\kern2pt}%
	\let\@fs@iftopcapt\iftrue}
\makeatother

\definecolor{citecolor}{HTML}{0071bc}
\usepackage[pagebackref=false,breaklinks=true,letterpaper=true,colorlinks,citecolor=citecolor,bookmarks=false]{hyperref}

\newcommand{\app}{\raise.17ex\hbox{$\scriptstyle\sim$}}

\newcommand{\apbbox}[1]{AP$^\text{bb}_\text{#1}$}
\newcommand{\apmask}[1]{AP$^\text{mk}_\text{#1}$}
\newcommand{\apkp}[1]{AP$^\text{kp}_\text{#1}$}

\newcolumntype{x}[1]{>{\centering\arraybackslash}p{#1pt}}
\newcolumntype{y}[1]{>{\raggedright\arraybackslash}p{#1pt}}
\newcolumntype{z}[1]{>{\raggedleft\arraybackslash}p{#1pt}}
\newlength\savewidth\newcommand\shline{\noalign{\global\savewidth\arrayrulewidth
		\global\arrayrulewidth 1pt}\hline\noalign{\global\arrayrulewidth\savewidth}}
\newcommand{\tablestyle}[2]{\setlength{\tabcolsep}{#1}\renewcommand{\arraystretch}{#2}\centering\footnotesize}

\makeatletter\renewcommand\subsubsection{\@startsection{subsubsection}{4}{\z@}
	{.5em \@plus1ex \@minus.2ex}{-.5em}{\normalfont\normalsize\bfseries}}\makeatother

\def\x{\times}

\usepackage{textpos}  

\begin{document}
	\title{Deeply Unsupervised Patch Re-Identification for Pre-training Object Detectors}
	\author{
		Jian Ding,
		Enze Xie, 
		Hang Xu,
		Chenhan Jiang,
		Zhenguo Li, 
		Ping Luo, 
		Gui-Song Xia \\
				\IEEEcompsocitemizethanks{
			\IEEEcompsocthanksitem This work was supported by National Nature Science Foundation of China under the grants 61922065, 41820104006 and 61871299. This work was also partially supported by the General Research Fund of HK No.27208720 and No.17212120. 
			\IEEEcompsocthanksitem J. Ding is with the State Key Lab. LIESMARS, and also School of Computer Scienc Wuhan University, Wuhan, 430079, China.  Email: jian.ding@whu.edu.cn.
			\IEEEcompsocthanksitem H. Xu, C. Jiang and Z. Li are with the Huawei Noah’s Ark Lab. Email: xbjxh@live.com, jchcyan@gmail.com, li.zhenguo@huawei.com.
			\IEEEcompsocthanksitem E. Xie and P. Luo are with the University of Hong Kong. Email: xieenze@hku.hk and pluo@cs.hku.hk.
			\IEEEcompsocthanksitem G.-S. Xia is with the National Engineering Research Center for Multimedia Software, School of Computer Science and Institute of Artificial Intelligence, and also the State Key Lab. LIESMARS, Wuhan University, Wuhan, 430072, China. Email: {guisong.xia}@whu.edu.cn.
			\IEEEcompsocthanksitem J. Ding and E. Xie are equally contributed to this work.
			\IEEEcompsocthanksitem Corresponding author: Gui-Song Xia (guisong.xia@whu.edu.cn.) 
		}
	}
	\IEEEtitleabstractindextext{%
		\justify
		\begin{abstract}
Unsupervised pre-training aims at learning transferable features that are beneficial for downstream tasks. However, most state-of-the-art unsupervised methods concentrate on learning \textit{global} representations for \textit{image-level} classification tasks instead of discriminative local region representations, which limits their transferability to \textit{region-level} downstream tasks, such as object detection. To improve the transferability of pre-trained features to object detection, we present \textit{Deeply Unsupervised Patch Re-ID} (DUPR), a simple yet effective method for unsupervised visual representation learning. The \textit{patch Re-ID} task treats individual patch as a pseudo-identity and contrastively learns its correspondence in two views, enabling us to obtain \textit{discriminative local features} for object detection. Then the proposed \textit{patch Re-ID} is performed in a \textit{deeply unsupervised} manner, appealing to object detection, which usually requires multi-level feature maps. Extensive experiments demonstrate that DUPR outperforms state-of-the-art unsupervised pre-trainings and even the ImageNet supervised pre-training on various downstream tasks related to object detection. 
		\end{abstract}		
	}
	\maketitle
	
	\IEEEraisesectionheading{
		\section{Introduction}
	}	

Pre-training then fine-tuning has been a widely used paradigm when approaching computer vision problems with deep models~\cite{r-cnn,FCN,maskrcnn,simonyan2014two}. In recent years, fine-tuning in object detection tasks has been dominated by the ImageNet supervised pre-training~\cite{r-cnn,maskrcnn,ren2015faster,lin2017focal}.
However, there exists misalignment between \textit{image-level} classification pre-training and object detection tasks, which make \textit{region-level} predictions. One solution to eliminate the misalignment~\cite{shao2019objects365} is to pre-train representations directly on a large-scale and high-quality object detection dataset; but annotations of such datasets are time-consuming, laborious, and even hard to obtain for some areas.

Alternatively, unsupervised learning aims at pre-training representations without human annotations, which allows us to pre-train representations with plenty of unlabeled data for free~\cite{Moco,deepcluster,bigbiggan,goodfellow2014generative,zhuang2019local,hinton1994autoencoders}. 
Among them, contrastive learning methods~\cite{Moco,SimCLR,instdis} have achieved comparable performance compared to ImageNet supervised pre-training in many downstream tasks, such as image classification, object detection, and semantic segmentation. 
Contrastive learning can learn \textit{view-invariant} representations by maximizing the similarity between positive pairs of views while minimizing the similarity between negative pairs.

\begin{figure}[t]
	\centering
	\vspace{-4mm}
	\includegraphics[width=0.92\linewidth]{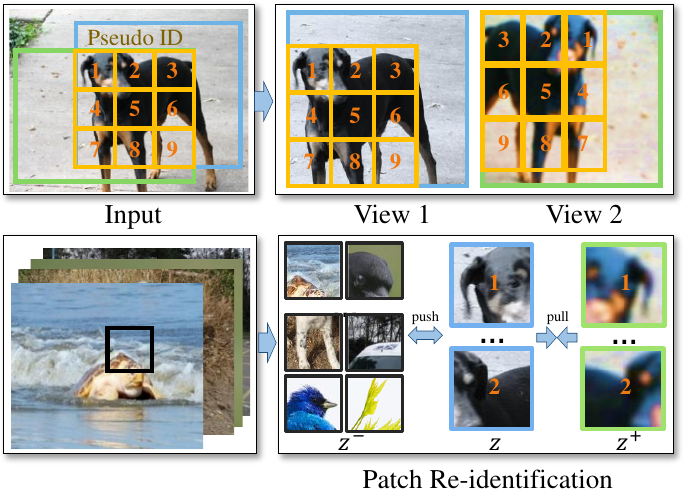}
	\vspace{-.1em}
	\caption{ Illustration of unsupervised pre-training for object detection by patch Re-ID, which follows the merits of person Re-ID that matches a human identity between cameras. In patch Re-ID, each patch within the intersection area (orange bounding box) is treated as a pseudo-identity. The unsupervised features are learned by matching the same patch identity (denoted by the same number) in two views generated by different transformations from the input image. In contrast to the previous global view based contrastive learning methods, which learn \textit{globally spatial invariant} representations, the patch Re-ID learns \textit{spatial sensitive} representations. Therefore, the translation, aspect ratio, and scale of objects in the input image will produce meaningful responses in the feature maps for describing the location of objects inside an image, which is beneficial for object detection. 
	}
	\label{fig:position_sentitive}
	\vspace{-4mm}
\end{figure}

Recent state-of-the-art contrastive learning methods take global views from the same image as positive pairs and views from different images as negative pairs.
The well-designed augmentations~\cite{SimCLR,infomin}, including random resized crop, are used to generate these views. Thus \textit{globally spatial invariant} representations will be learned, which are beneficial for classification. 
For example, classification models should predict the same category "dog" for two views generated from the same image, containing a dog at the bottom-left and top-left (see Fig.~\ref{fig:position_sentitive}).
So, these two views containing the same object but at different locations should have similar global representations. 
The representations learned by contrasting global views can encode
much information related to category and largely improve the performance on ImageNet linear evaluation, approaching the accuracy of supervised classification~\cite{SimCLR,Mocov2,simsiam,byol}.

However, there is a gap between \textit{pre-training global representations} and \textit{region-level} downstream tasks, such as object detection. Being different from \textit{image-level} classification, which predicts the category for an entire image by a \textit{globally spatial invariant} feature, object detection is a \textit{region-level} task which predicts categories and regression targets for multiple regions by the region features.
The region features at different locations should be discriminative since the prediction targets for these regions are different. For example, regions of interest (RoIs) are assigned with foreground or background categories. 
Also, foreground RoIs overlapping objects at different positions should predict different regression targets.
Based on the above reasons, features at different locations in a feature map should be mapped to far apart points in \textit{local feature space} (see Fig.~\ref{fig:pipeline}).
Thus, the previous methods that only optimize the single feature after globally averaged pooling are problematic for object detection since they do not learn discriminative local representations in a feature map. So the previous methods with higher performance on ImageNet classification do not always lead to better transfer performance on object detection~\cite{Mocov2,byol,swav} (see also in Tab.~\ref{tab:detvscls}).

Moreover, previous works focus on learning discriminative features at the final layer ({\em \eg}, on the $32\x$ feature map)~\cite{Moco,SimCLR}. However, most deep learning based object detectors need to extract features from the multi-level representations (such as FPN~\cite{FPN} and PANet~\cite{panet}).
Thus, object detection requires discriminative features at different feature layers instead of only the final layer. 

To address the aforementioned problems, we propose to pre-train \textit{region-level} discriminative representations across multi-level feature maps for object detection by \textit{Deeply Unsupervised Patch Re-ID} (DUPR). The task of patch Re-ID is to match the corresponding patch identity (denoted by the same number in Fig.~\ref{fig:position_sentitive}) of two views, which is inspired by the person Re-ID that matches a human identity between cameras under different viewing conditions. By pre-training with patch Re-ID, the features of matched patches should be more similar than unmatched patches in \textit{local feature space} (see Fig.~\ref{fig:pipeline}), so the region representations at different locations in a feature map are discriminative and beneficial for \textit{region-level} tasks such as object detection. Furthermore, we propose a {\em deeply unsupervised training strategy} to learn multi-level representations. Specifically, we extract features from different intermediate layers to construct both \textit{image-level} and \textit{patch-level} contrastive loss. Our DUPR is independent of the detailed self-supervised learning framework. We simply adopt the MoCo framework~\cite{Moco,Mocov2} and InfoNCE~\cite{CPC} as loss function in this work, but patch Re-ID can also be used in other self-supervised learning frameworks~\cite{SimCLR,byol}. The whole pipeline is shown in Fig.~\ref{fig:pipeline}.

\begin{figure}[t!]\centering
\includegraphics[width=0.98\linewidth]{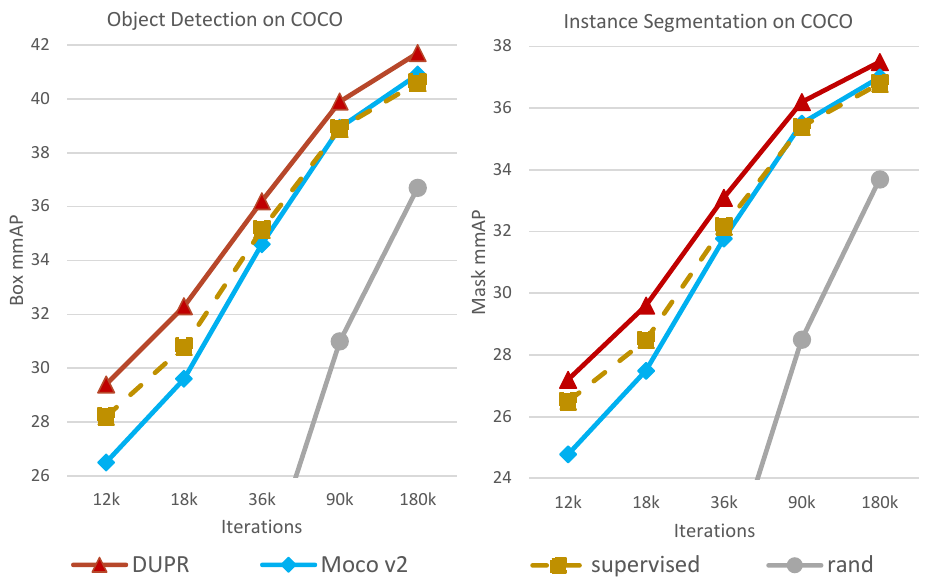}
\vspace{-0.5em}

		\caption{\textbf{Comparison with other pre-training methods on fine-tuning Mask R-CNN R-50-FPN on COCO}. The MoCo v2 baseline~\cite{Mocov2} and DUPR are pre-trained on the ImageNet-1M training set with 200 epochs. DUPR outperforms MoCo v2 and even the supervised counterpart at all iterations. (a) shows the box mmAP and (b) shows the mask mmAP. When fine-tuning Mask R-CNN with 12k iterations, DUPR outperforms the MoCo v2 by 2.9 points in box mAP and 2.4 points in mask mAP. 
	\label{fig:iteration_accuracy}}
\end{figure}

Our contributions can be summarized as follows:
\begin{itemize}
	\item We propose a self-supervised pretext task, named {\em patch Re-ID} that learns to match the same patch identity between two views to get \textit{region-level} discriminative feature maps, which is tailored for object detection.
	\item We present a deeply unsupervised training strategy to improve the transferability of pre-trained models to object detection, which requires extracting features from multi-level feature maps for prediction.
	\item Our DUPR pre-training outperforms other unsupervised and supervised pre-training counterparts when serving as the initialization for fine-tuning. For example, when fine-tuning Mask R-CNN R-50-FPN on MS COCO~\cite{coco}, DUPR outperforms MoCo v2~\cite{Mocov2} and supervised pre-training at all different iterations as shown in Fig.~\ref{fig:iteration_accuracy}. More importantly, it outperforms the strong baseline MoCo v2 when serving as initialization for fine-tuning in other location-sensitive tasks, such as VOC~\cite{PASCALVOC} object detection (+2.4 \apbbox{75}), Cityscapes~\cite{Cityscapes} semantic segmentation (+1.0 mIoU), and LVIS~\cite{lvis} instance segmentation (+1.0 \apmask{~}).
	 
\end{itemize}


\section{Related Work}
\subsection{Object Detection}
The task of object detection is to locate objects in images and classify their categories. Unlike the \textit{image-level}  classification prediction task, object detection is a \textit{region-level} prediction task:  it needs to regress the location and classify the category of an object simultaneously for each \textit{region}. 
In two-stage object detectors~\cite{maskrcnn,ren2015faster}, region features are extracted from the \textit{proposals} (generated by selective search~\cite{selectivesearch} or RPN~\cite{ren2015faster}) by RoI Align~\cite{maskrcnn}. For one-stage object detectors~\cite{lin2017focal,ssd,yolo,tian2019fcos}, region features are extracted from \textit{sliding windows}.

The prediction targets are different for different regions in an image. For each region, the \textit{classification branch} usually predicts the confidence score of categories (foreground and background in RPN~\cite{ren2015faster}, or categories of objects in one-stage detector~\cite{lin2017focal} and R-CNN~\cite{r-cnn}). 
A region of interest (RoI) is assigned to be a positive example if it has an intersection-over-union (IoU) overlap with a ground-truth box above a threshold; otherwise, it is assigned to be a negative example. Therefore, the category prediction target is sensitive to the change of regions (when the IoU changes from below the threshold to above the threshold, or from one object to another object with a different category). The \textit{localization branch} usually predicts the regression targets relative to the anchor (see definition of regression targets in ~\cite{ssd,r-cnn}), and different positive RoIs should predict different regression targets. For example, an RoI having high IoU overlap with an object should predict small regression targets, while an RoI having low IoU overlap with an object should predict large regression targets (see Fig.~\ref{fig:predictiontarget}).
So the local region representations at different locations should be mapped to far apart points in \textit{local feature space} for object detection. Besides, object detection is also the combination of region-level classification and localization. Features need to be sensitive to the location of a feature map for object localization while maintaining the strong semantic information for classification at the same time.

Object detection also requires \textit{multi-level} representations, as predictions are directly made by using multi-level features~\cite{ssd}, or the fusion of multi-level feature maps~\cite{FPN,panet}. For example, FPN~\cite{FPN} is a widely used structure in object detection to handle the scale variations, which combines low-resolution, semantically strong features with high-resolution, semantically weak features via top-down connections.
Thus, our work will focus on learning discriminative region-level features from multiple layers to pre-train representations, which are tailored for object detection.

\subsection{Pre-training for Object Detection}
R-CNN~\cite{r-cnn} has shown that ImageNet supervised pre-training followed by domain-specific fine-tuning on a small dataset is an effective paradigm for learning high-capacity representations. Pre-training for object detection largely improves the performance on small dataset~\cite{r-cnn,rethinking} and also speeds up the convergence of object detectors~\cite{rethinking}.
However, ImageNet supervised pre-training is weak at localization and helps less if the downstream task is localization-sensitive~\cite{r-cnn,rethinking}.
To obtain better pre-trained representations for object detection, Objects365~\cite{shao2019objects365} is proposed. Pre-training on this large-scale and high-quality object detection dataset can significantly surpass the ImageNet supervised pre-training in convergence speed and mAP. Since annotations of object detection are expensive, weakly supervised pre-training~\cite{weakpretrain} has been explored for object detection. However, the weakly supervised pre-training pipeline~\cite{weakpretrain} is complicated and still needs annotations. In contrast to these works, our paper presents a method for unsupervised pre-training for object detection, which 
has rarely been studied before. 

\subsection{Self-supervised Visual Representation Learning}
Self-supervised visual representation learning leverages input data itself as supervision via pretext tasks. After pre-training, features are transferred to downstream tasks. Early pretext tasks for self-supervised representation learning include rotation prediction~\cite{rotation}, relative location prediction~\cite{relativeloc} and jigsaw~\cite{jigsaw}, etc. 
These hand-crafted pretext tasks achieve promising results but still have a large gap compared to ImageNet supervised pre-training. 
Recently, the most successful methods in self-supervised learning are contrastive learning~\cite{hadsell2006dimensionality,Moco,CPC,SimCLR,infomin,CMC,CPCv2,simsiam,byol,swav} via instance discrimination pretext task~\cite{instdis,Moco,exemplar}. The core idea of contrastive learning is to pull together positive view pairs while pushing apart negative view pairs.

The success of contrastive learning relates to learning the invariant representations to a family of similar views (positive pairs). The selection of positive pairs and data augmentations on views is important, and varies in different methods. For example, CPC~\cite{CPC} and CPC v2~\cite{CPCv2} take the context and future as positive pairs. Deep infomax~\cite{deepinfomax} and AMIDIM~\cite{aidim} take the global and local features as two positive pairs. 
MoCo~\cite{Moco} and SimCLR~\cite{SimCLR} adopt the instance discrimination task~\cite{exemplar,instdis} which takes the randomly augmented global views from the same image as two positive pairs. SimCLR has also studied many data augmentation strategies to generate views.
SwAV has additionally proposed a novel multi-crop data augmentation, which increases the number of local views and maximizes the similarity between global and local views. The influence of augmentations and view selection for different downstream tasks has been studied in ~\cite{infomin,looc}. Different downstream tasks have different optimal selections of positive view pairs.
\revision{It has been proven in~\cite{infomin} that the optimal views should share the minimum information necessary to perform well at the downstream tasks.}
However, InfoMin~\cite{infomin} still takes the classification as the downstream task and designs augmentations to improve the classification performance.

\blank{Most of the previous methods maximize the similarity between spatial misaligned views and focus on learning \textit{globally} \textit{spatial invariant} representations for \textit{image-level} classification, although the details are different. 
\revision{However, the better classification accuracy of linear probing does not always lead to better transfer performance to object detection.}
For example, SwAV~\cite{swav} and BYOL~\cite{byol} are much higher than MoCo v2~\cite{Mocov2} in ImageNet linear probing, but lower in the transferring performance on object detection (see Tab.~\ref{tab:detvscls}).}
\blank{Being different form these works, our approach maximizes the similarity between the spatially consistent local and local views, and aims at learning \textit{spatial sensitive}, \textit{multi-level}, \textit{entire feature maps} for \textit{region-level} tasks such as object detection.} In the way of extracting local views, most of the previous methods send extra patches from the initial image to the network, such as SwAV~\cite{swav} and CPC~\cite{CPC,CPCv2}. In contrast, our method directly extracts the local representation from the feature map, which is more efficient.

\begin{figure*}[t]
	\centering
	\vspace{-2mm}
	\includegraphics[width=0.98\linewidth]{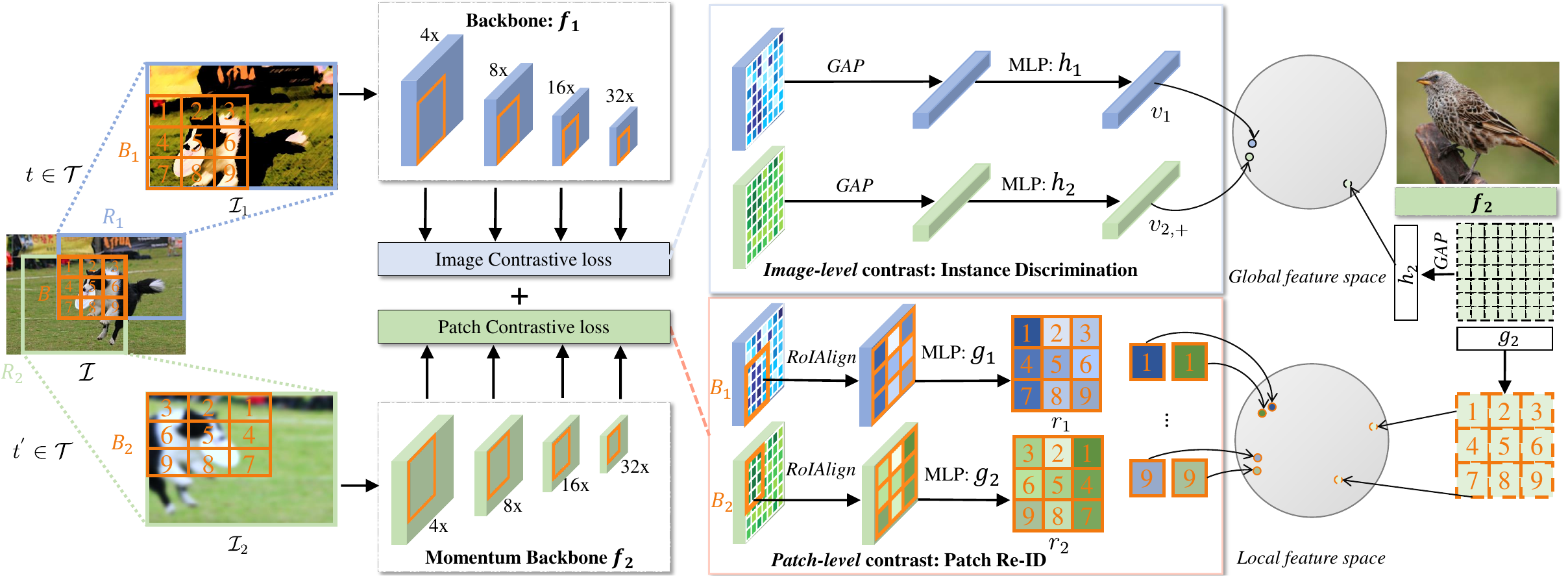}
	\caption{\textbf{The pipeline of Deeply Unsupervised Patch Re-ID (DUPR).} The original image $\mathcal{I}$ is augmented into two views $\mathcal{I}_1$ and $\mathcal{I}_2$. The blue and green bounding boxes correspond to the original area of $\mathcal{I}_1$ and $\mathcal{I}_2$, respectively. The orange bounding box is the intersection area. $\mathcal{I}_1$ and $\mathcal{I}_2$ are passed to the backbone and momentum backbone, respectively. We take four feature levels to construct the contrastive loss. Each feature level includes an \textit{image-level} and a \textit{patch-level} contrastive loss. The \textit{image-level} contrastive learning aims at mapping similar global views to the nearby points in \textit{global feature space} but has no constraint on the local representations. \textit{Patch-level} contrastive learning aims at mapping the corresponding \textit{local features} of patch identity (denoted by the number in this figure) to nearby points in \textit{local feature space}. 
	To obtain the corresponding patch identity, the intersection area is numerically calculated then RoI Align is applied to the intersection area to extract region features with the shape of $(C, S\times S)$. Flipping is also applied to region features if the input image is flipped. The similarity between matched patch features are maximized by the \textit{patch-level} contrastive loss, making representation discriminative at different locations. The parallel \textit{image-level} and \textit{patch-level} contrastive learning makes the representation both semantic strong and location-sensitive, tailored for object detection.
		\label{fig:pipeline}}
	\vspace{-1.em}
\end{figure*}
\subsection{Unsupervised Dense Representation Learning}

\blank{There are some early works related to pixel or region-level representation learning. The auto-encoders~\cite{zhang2016colorful,denoiseautoencoder,contextencoder} are trained by generating or modeling pixels in the input space. 
However, pixel-level generation is computationally expensive and needs extra heavy decoders which are not used by the downstream tasks. And the cycle-consistency of time in video sequences~\cite{timecycle,unsuptrack} use self-supervised tracking as a pretext task and learns the pixel-level correspondence in the video sequences. However, these works aim at directly being deployed in visual correspondence tasks in videos without fine-tuning, rather than being transferred to other downstream tasks. 
}

\blank{Recently, VADeR~\cite{VADeR} has explored the \textit{pixel-level contrastive learning} for transferring to multiple dense prediction tasks. But the VADeR requires the initialization of MoCo~\cite{Moco} and only optimizes the local representation; our model is trained from scratch, and optimizes both global and local representation. The VADeR~\cite{VADeR} also does not perform well at object detection.

Being concurrent to our work, \revision{there are several self-supervised learning methods~\cite{instloc,selfemd,detco,pixpro,densecl} targeting at object detection and semantic segmentation. InstLoc~\cite{instloc} crops two \textit{spatially misaligned} patches and pastes them on two background images to form two positive views for contrastive learning, which is quite different from our method. DetCo~\cite{detco} focuses on the trade-off between classification and object detection tasks, and proposes to use global and local contrastive learning. Self-EMD~\cite{selfemd}, DenseCL~\cite{densecl}, and PixPro~\cite{pixpro} are three dense self-supervised learning methods. Self-EMD~\cite{selfemd} and DenseCL~\cite{densecl} find the positive pairs of patch features by the similarities between patch features, which are unstable if the initialization is not good. Different from them, DUPR finds the positive patches by their locations in the original images, which is more accurate and stable.} 
\begin{figure}[t!]
	\centering
	\includegraphics[width=.9\linewidth]{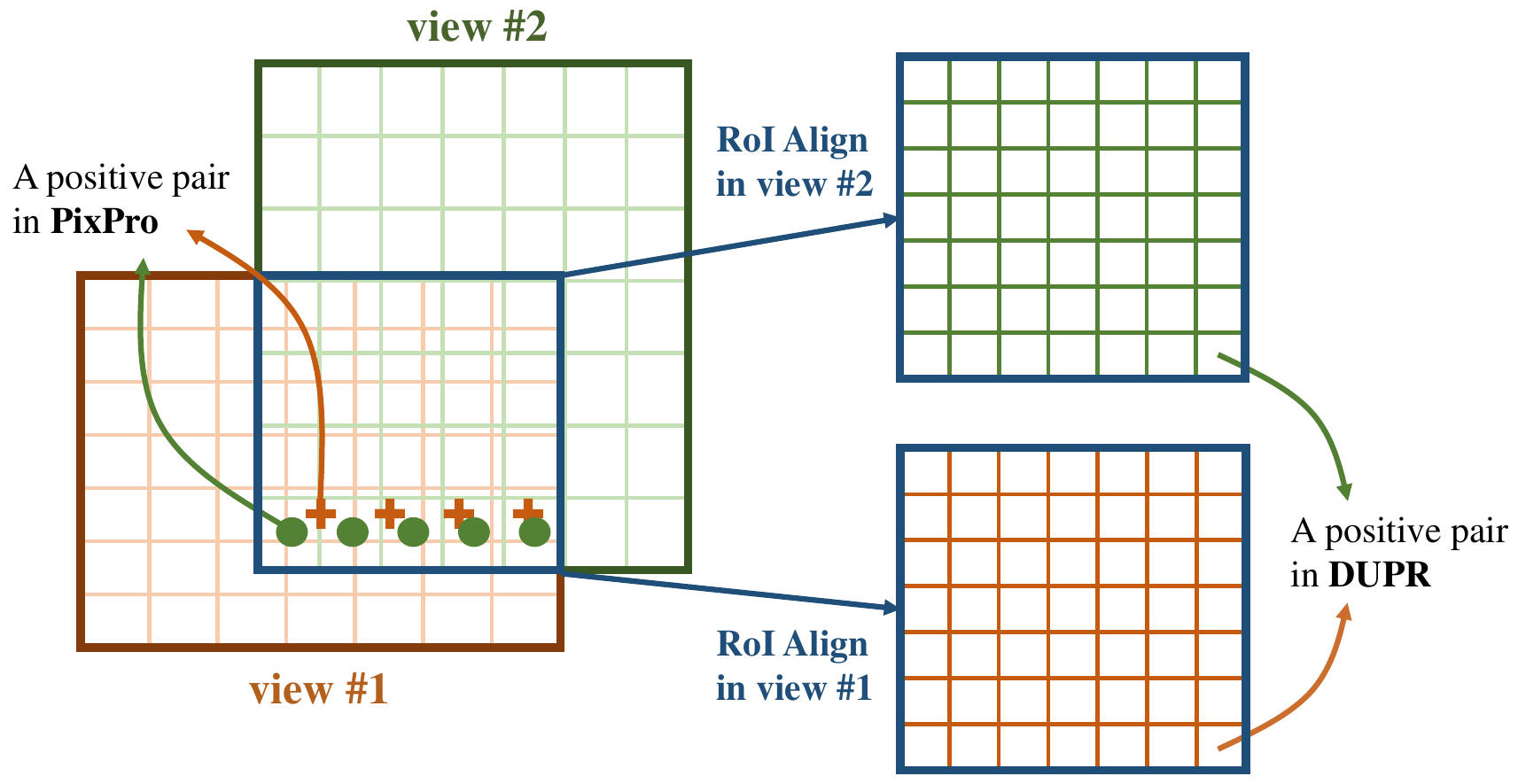}
	\caption{\textbf{Comparison between PixPro~\cite{pixpro} and DUPR in the construction of positive pairs.} PixPro~\cite{pixpro} finds positive pairs in two feature maps with \textit{spatially misaligned grids}: two patches are considered as positive pairs if the distance between their centers is below a threshold. DUPR generates two feature maps with \textit{spatially aligned grids}, and can naturally obtain the positive pairs in the \textit{aligned} feature maps.
		\label{fig:pixcontrast}}
	\vspace{-1.em}
\end{figure}

\revision{PixPro~\cite{pixpro} is mostly related to our work since it also finds the positive pairs of patches by locations in the original images; but we are different in the details for the construction of positive pairs, as shown in Fig.~\ref{fig:pixcontrast}. PixPro~\cite{pixpro} finds the matched patches in two \textit{spatially misaligned grids}, and a threshold of the distance between misaligned patches is introduced. In contrast, we apply RoI Align~\cite{maskrcnn} on the intersection of two views and generate feature maps with \textit{spatially aligned grids}.
So we can naturally obtain the matched patches, and do not need an extra hyperparameter. PixPro~\cite{pixpro} also introduces a pixel propagation module (PPM) to further improve the performance. Being \textit{simpler} than PixPro~\cite{pixpro}, our method performs \textit{slightly better}, as shown in Tab.~\ref{tab:voc_detection} and Tab.~\ref{tab:coco}. } 
}
\section{Method}
\subsection{Preliminary: Contrastive Learning}	
\blank{The main idea of contrastive learning is to pull together positive views while pushing apart negative views. Take MoCo~\cite{Moco} as an example, suppose $\mathcal{I}$ is the original image,
$\mathcal{I}_{1}$ and $\mathcal{I}_{2}$ can be considered as two views of the same image with different augmentations.
Denote $v_{1}$ and $v_{2,+}$ to be the normalized embeddings of $\mathcal{I}_{1}$ and $\mathcal{I}_{2}$. The target of contrastive learning is to pull together positive pairs ($v_{1}$, $v_{2,+}$) while pushing apart negative pairs ($v_{1}$, $v_{2,j}$). The conventional learning objective is the InfoNCE~\cite{CPC} loss:
\begin{equation}
	\small
	\mathcal{L}_{v_{1},v_{2,+}} = -\log \frac{\exp(v_{1}{\cdot}v_{2,+} / \tau)}{\sum_{j=0}^{K}\exp(v_{1}{\cdot}v_{2,j}  / \tau)}.
	\label{eq:infonce}
\end{equation}
Here $\tau$ is a temperature hyper-parameter. $v_{1} {\cdot} v_{2,j}$ is the cosine similarity to measure the distance between two image features.  
It can be considered as a non-parametric softmax-based classifier to identify $v_{2,+}$ as $v_{1}$. 
} 

\blank{Selecting positive pairs is important [15], [41] for contrastive learning since they will learn invariant representation against the transformation applied on positive pairs. What kind of transformation should the representation be invariant to is determined by the downstream tasks, and varies across different downstream tasks~\cite{infomin,looc}. For example, suppose the downstream task is classification; in such case, the representation \textit{should be invariant} to the location of an object inside an image since the change of locations does not change the semantic category. 
And, suppose the downstream task is to predict the location of an object. In this case, the representation \textit{should not be invariant} to the locations. But other factors (\eg, category and light condition) are irrelevant information, to which the representation should be invariant. 

In the previous works, $v_{1}$ and $v_{2,+}$ are pairs of global-local features~\cite{aidim,deepinfomax} or spatial misaligned global-global features~\cite{Moco,SimCLR}. On the one hand, these global averaged features will lose spatial information. On the other hand, the positive pairs are not spatially aligned. Thus, these global view based methods tend to learn \textit{globally spatial invariant feature}, which is suitable for \textit{image-level} classification but not for \textit{region-level} and \textit{location-sensitive} tasks such as object detection. In contrast to the previous methods, our method can be considered as spatially aligned local-local views selection, where the representations are sensitive to the locations of objects, while invariant to other factors.}

\subsection{Deeply Unsupervised Patch Re-ID}\label{sec:DUPR}
The pipeline of DUPR is shown in Fig.~\ref{fig:pipeline}, which consists of \textit{patch-level} and \textit{image-level} contrastive learning in parallel across multi-feature levels. The \textit{patch-level} contrastive learning directly optimizes an entire feature map before averaged pooling and maximizes the similarity between matched patches (denoted by the same number in Fig.~\ref{fig:pipeline}) to strengthen the spatial information for localization. Since object detection is the combination of localization and classification, we also include the \textit{image-level} contrastive learning to strengthen the semantic information for classification.
For the contrastive learning framework, we simply choose the MoCo v2~\cite{Mocov2} as our strong baseline, although other contrastive learning frameworks are also possible. 

Finally, we add the \textit{patch-level} and \textit{image-level} contrastive loss to multi-feature levels, as most object detectors need a multi-level representation, such as FPN~\cite{FPN} and PANet~\cite{panet}. The overall loss is defined as:
\begin{equation}
	\small
	\mathcal{L} = \sum_{m=0}^{M}\mathcal{\alpha}_{m}\mathcal{L}^{(m)}_{image} + \sum_{m=0}^{M}\mathcal{\beta}_{m}\mathcal{L}^{(m)}_{patch}
	\label{eq:loss_all}
\end{equation}
where $M$ is the number of feature maps. $\mathcal{L}^{(m)}_{image}$ and $\mathcal{L}^{(m)}_{patch}$ are the image and patch contrastive loss for feature map of $m$-th level. $\mathcal{\alpha}_m$ and $\mathcal{\beta}_m$ are the weights to balance the importance of different levels. 
We will describe the details of \textit{patch-level} and \textit{image-level} contrastive learning across multi-level feature maps in the following.

\subsubsection{Patch-Level Contrastive Loss}\label{patch-reid}
\begin{figure}[t!]
	\centering
	\includegraphics[width=.97\linewidth]{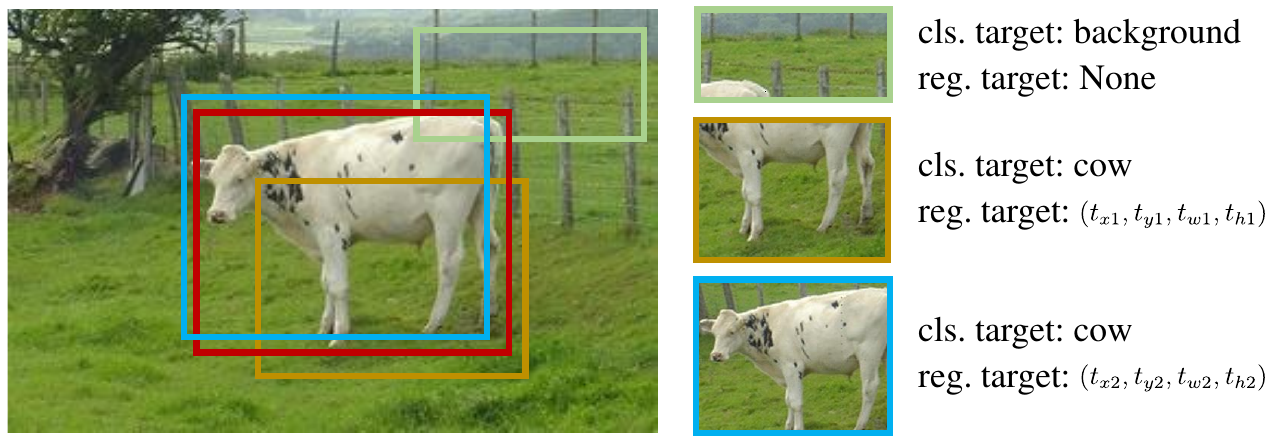}
	\vspace{-1mm}
	\caption{\textbf{The prediction targets varies across different regions}. The cls. denotes the classification target while the reg. denotes the regression target. The $(t_x, t_y, t_w, t_h)$ is an instantiation of regression target, which follows the definition in~\cite{r-cnn}, 
	but it can also be any other kinds of regression targets, such as~\cite{fcos}. In this figure, the absolute value of $(t_{x_1}, t_{y_1}, t_{w_1}, t_{h_1})$ is larger than $(t_{x_2}, t_{y_2}, t_{w_2}, t_{h_2})$.
		\label{fig:predictiontarget}}
	\vspace{-1.em}
\end{figure}

\blank{The previous methods optimize global representations and do not learn discriminative local representations. However, for location-sensitive tasks, features of different patches in a feature map should be different since their prediction targets are different. 
For example, different regions may represent different categories of objects or the background. The regression targets for different regions are also different (see Fig.~\ref{fig:predictiontarget} for illustration). So the local features of the matched patches should be mapped to the nearby points in \textit{local feature space}, while local features of different patches should be mapped to far apart points in \textit{local feature space} as shown in Fig.~\ref{fig:pipeline}.}

To learn \textit{patch-level} discriminative representations, we design the \textit{patch Re-ID pretext task} to match the same patch identities of two views (which are denoted by the same number in Fig.~\ref{fig:pipeline}).
First, we sample two augmentations ($t\in \mathcal{T}$ and $t^{'}\in \mathcal{T}$) from the same family of augmentations\revision{: each augmentation is a composition of multiple transformations (\eg, cropping, resizing, flipping, and color distortion.)} 
Then we apply augmentations to input image $\mathcal{I}$ and get two views: $\mathcal{I}_1=t(\mathcal{I})$ and $\mathcal{I}_2=t^{'}(\mathcal{I})$. The corresponding rectangular regions of $\mathcal{I}_1$ and $\mathcal{I}_2$ in the original image $\mathcal{I}$ are recorded and denoted as:
\begin{equation}
	\small
	\begin{split}
	R_1 &= (tl_x^{(1)}, tl_y^{(1)}, br_x^{(1)}, br_y^{(1)})\\
	R_2 &= (tl_x^{(2)}, tl_y^{(2)}, br_x^{(2)}, br_y^{(2)})
	\end{split}    
\end{equation}
where the $(tl_x, tl_y)$ denotes the top-left vertex and $(br_x, br_y)$ denotes the bottom-right vertex of a rectangular region. Then we can calculate the intersection area $\mathcal{B}=(tl_x^{(B)}, tl_y^{(B)}, br_x^{(B)}, br_y^{(B)})$ from $R_1$ and $R_2$ as:
\begin{equation}
	\begin{split}
	tl_x^{(B)} &= \max(tl_x^{(1)}, tl_x^{(2)})\\
	tl_y^{(B)} &= \max(tl_y^{(1)}, tl_y^{(2)})\\
	br_x^{(B)} &= \min(br_x^{(1)}, br_x^{(2)})\\
	br_y^{(B)} &= \min(br_y^{(1)}, br_y^{(2)})\\
	\end{split}
\end{equation}
 The intersection $\mathcal{B}$ in the coordinate system
 of $\mathcal{I}_{1}$ and $\mathcal{I}_{2}$ can be obtained by $\mathcal{B}_1=T_1(\mathcal{B})$ and $\mathcal{B}_2=T_2(\mathcal{B})$, where the $T_1$ and $T_2$ are coordinate transformations from $\mathcal{I}$ to $\mathcal{I}_1$ and $\mathcal{I}$ to $\mathcal{I}_2$, respectively.
 \blank{Instead of using a global averaged feature, which loses spatial information, we split $B_1$ and $B_2$ of the two views into $S\times S$ patches and maximize the similarity between the corresponding patch features (denoted by the same number).} 
For the detailed implementation of $m$-th feature map, we apply RoI Align~\cite{maskrcnn} to extract region feature followed by a pixel-wise MLP layer which is implemented by a $1\times 1$ convolution as:
\begin{align}
	\small
	r^{(m)}_{1} &= g^{(m)}_1(RoIAlign(f^{(m)}_1(\mathcal{I}_{1}), \mathcal{B}_{1}))
	\label{eq:roialign_q} \\
	r^{(m)}_{2} &= g^{(m)}_2(RoIAlign(f^{(m)}_2(\mathcal{I}_{2}), \mathcal{B}_{2}))
	\label{eq:roialign_k}
\end{align}
where $g_1^{(m)}$ and $g_2^{(m)}$ are MLP layer and momentum MLP layer, respectively. 
$r^{(m)}_{1}$ and $r^{(m)}_{2}$ are region features of a fixed shape $(C, S \times S)$. 
Then $r^{(m)}_{1,p}$ and $r^{(m)}_{2,p}$ is a positive pair of normalized feature vectors, where the subscript $p\in [0,S\times S)$ denotes the position in the intersection area. Although our patch Re-ID pretext task is independent of loss functions, we simply adopt the InfoNCE~\cite{CPC} loss and follow MoCo v2~\cite{Mocov2} to use a dynamic memory bank to store features from momentum updated encoder.  
We construct the \textit{patch-level} contrastive loss for the $m$-th feature map as:
\begin{equation}
	\scriptsize
	\mathcal{L}^{(m)}_{patch} = -\sum_{p}\log \frac{\exp(r_{1,p}^{(m)}{\cdot}r_{2,p}^{(m)}/\tau)}{\exp(r_{1,p}^{(m)}{\cdot}r_{2,p}^{(m)}/\tau) + \sum_{t=1}^{K}\exp(r_{1,p}^{(m)}{\cdot}r_t^{(m)}  / \tau)}
	\label{eq:singlepatchnce}
\end{equation}
where $\{r_t^{(m)}\}_{t=1,...K}$ are patch features of other images from memory bank. $m$ denotes the index of feature map.
By minimizing the patch-level contrastive loss, our encoder can learn patch-wise matching of identity between two views. Such matching ability results in spatially sensitive feature maps, which facilitates object detection. Note that if there is no overlap between $R_1$ and $R_2$, we will ignore the \textit{patch-level} contrastive loss. The probability of such a case is very low \revision{(\eg, 51 out of 1472 positive pairs), so it has not much effect on the results.}

\subsubsection{Image-Level Contrastive Loss}
We also optimize the image-level contrastive loss as it is important to improve the classification ability, which is also required by object detection. 
Denote $v_{1}^{(m)}=h_1(GAP(f_1^{(m)}(\mathcal{I}_{1})))$ and $v_{2,+}^{(m)}=h_2(GAP(f_2^{(m)}(\mathcal{I}_{2})))$ to be the normalized image features of positive pairs. For simplicity, we neglect the notation of normalization. The \textit{image-level} contrastive loss for the $m$-th feature map can be written as: 
\begin{equation}
	\small
	\mathcal{L}^{(m)}_{image} = -\log \frac{\exp(v_{1}^{(m)}{\cdot}v_{2,+}^{(m)} / \tau)}{\sum_{j=0}^{K}\exp(v_{1}^{(m)}{\cdot}v_{2,j}^{(m)}  / \tau)}
	\label{eq:singlegap}
\end{equation}

\subsubsection{Implementation Details}
We use unlabelled ImageNet to pre-train our models for our experiments. \blank{For the ablation experiments, we follow the data augmentation settings in~\cite{Mocov2}. For the main experiments, we add the Rand-Augmentation~\cite{randaugment} following~\cite{infomin}.} 
We choose ResNet 50~\cite{resnet} as our backbone and extract multi-level features from conv2\_x, conv3\_x, conv4\_x and conv5\_x. The stride of each feature map is $\{4\x, 8\x, 16\x, 32\x\}$, respectively. By default, we set $\alpha_{0:3}=(0.1,0.4,0.7,1.0)$ and $\beta_{0:3}=(0,0,1,1)$ for Eq.~\eqref{eq:loss_all}. The RoI size $S$ of patch features on conv5\_x and conv4\_x are 14 and 7.
$\tau$ is 0.2 for ablation experiments and 0.15 for main experiments. Unless other specified, we train with a batch size of 256 for 200 epochs. We use a learning rate of 0.06 with a cosine decay schedule. 

\revision{We maintain a unique memory bank for each image-level and patch-level contrastive loss in Eq.~\eqref{eq:loss_all}. For $\mathcal{L}_{patch}^{(m)}$, the memory bank stores the patch features of the $m$-th feature map of other images. For $\mathcal{L}_{image}^{(m)}$, the memory bank stores the image features of the $m$-th feature map of other images.}
We store 65536 keys for each memory bank. For patch features, there are $S\times S$ features for a single level on a single image, where $S=7$ on conv5\_x and $S=14$ on conv4\_x. In a batch with 256 images, there are $256\times S\times S$ patch features. Since most patch features from a single image are similar, we sample \revision{32 patch features per batch}, when we enqueue and dequeue a batch of patch features. 

All the algorithms of object detection, instance segmentation, and semantic segmentation are implemented in detectron2~\footnote{\url{https://github.com/facebookresearch/detectron2}}. For Mask R-CNN R-50-FPN, Mask R-CNN R-50-C4, and Faster R-CNN-C4, we follow the settings in MoCo~\cite{Moco}.  For RetinaNet R-50-FPN, we also add an extra normalization layer similar to Mask R-CNN R-50-FPN in MoCo~\cite{Moco}. 
The pre-trained weights of PIRL~\cite{misra2020self}, InsDis~\cite{instdis} are downloaded from Pycontrast~\footnote{\url{https://github.com/HobbitLong/PyContrast}}, while the pre-trained weights of SwAV is downloaded from the official code~\footnote{\url{https://github.com/facebookresearch/swav}}.

\section{Experimental Analysis}
We evaluate our DUPR and compare it with recent state-of-the-art unsupervised and supervised counterparts in various object detection related downstream tasks. The results of PASCAL VOC~\cite{PASCALVOC} object detection are reported in Sec.~\ref{sec:vocdetection}. The results of COCO~\cite{coco} object detection and instance segmentation are presented in Sec.~\ref{sec:coco_results}. Other localization-sensitive tasks (COCO keypoint detection, Cityscapes~\cite{Cityscapes} semantic segmentation, and instance segmentation, and LVIS instance segmentation~\cite{lvis}) are presented in Sec.~\ref{sec:othertask}. The comparisons with other methods in object detection v.s. classification performance are reported in Sec.~\ref{sec:cls_acc}.
Then we analyze some ablation experiments in Sec.~\ref{sec:experiments}, and give the visualization of features in Sec.~\ref{sec:visualization}. 

\definecolor{Gray}{gray}{0.5}

\newcommand{\randinit}{\tablestyle{1pt}{1} \begin{tabular}{z{21}y{26}} \multicolumn{2}{c}{\demph{random init.}} \end{tabular}}
\newcommand{\mocoimgnet}{\tablestyle{0pt}{1} \begin{tabular}{z{21}y{26}} \textbf{MoCo} & ~~IN-1M \end{tabular}}
\newcommand{\mocoins}{\tablestyle{0pt}{1} \begin{tabular}{z{21}y{26}} \textbf{MoCo} & ~~IG-1B \end{tabular}}
\newcommand{\supimgnet}{\tablestyle{0pt}{1} \begin{tabular}{z{21}y{26}} super. & ~~IN-1M \end{tabular}}
\newcommand{\levelablationone}{\tablestyle{0pt}{1} \begin{tabular}{z{21}y{26}} (0,0,0,1) \end{tabular}}
\newcommand{\levelablationtwo}{\tablestyle{0pt}{1} \begin{tabular}{z{21}y{26}} (0,0,1,1) \end{tabular}}
\newcommand{\levelablationthree}{\tablestyle{0pt}{1} \begin{tabular}{z{21}y{26}} (1,1,1,1) \end{tabular}}
\newcommand{\levelablationfour}{\tablestyle{0pt}{1} \begin{tabular}{z{21}y{26}} (0.1,0.4,0.7,1) \end{tabular}}

\newcommand{\demph}[1]{\textcolor{Gray}{#1}}
\newcommand{\std}[1]{{\fontsize{5pt}{1em}\selectfont ~~$_\pm$$_{\text{#1}}$}}

\definecolor{Highlight}{HTML}{39b54a}  

\renewcommand{\hl}[1]{\textcolor{Highlight}{#1}}

\renewcommand{\ll}[1]{\textcolor{red}{#1}}

\newcommand{\res}[3]{
	\tablestyle{1pt}{1}
	\begin{tabular}{z{16}y{18}}
		{#1} &
		\fontsize{7.5pt}{1em}\selectfont{~(${#2}${#3})}
\end{tabular}}

\newcommand{\reshl}[3]{
	\tablestyle{1pt}{1} 
	\begin{tabular}{z{16}y{18}}
		{#1} &
		\fontsize{7.5pt}{1em}\selectfont{~\hl{(${#2}$\textbf{#3})}}
\end{tabular}}

\newcommand{\resll}[3]{
	\tablestyle{1pt}{1} 
	\begin{tabular}{z{16}y{18}}
		{#1} &
		\fontsize{7.5pt}{1em}\selectfont{~\ll{(${#2}$\textbf{#3})}}
\end{tabular}}

\newcommand{\resrand}[2]{\tablestyle{1pt}{1} \begin{tabular}{z{16}y{18}} \demph{#1} & {} \end{tabular}}
\newcommand{\ressup}[2]{\tablestyle{1pt}{1} \begin{tabular}{z{16}y{18}} {#1} & {} \end{tabular}}

\begin{table}[t!]
	\caption{\textbf{Object detection fine-tuned on PASCAL VOC.}
	All the methods are pre-trained on ImageNet-1M. We fine-tune the Faster R-CNN R-50-C4 on Pascal VOC \texttt{trainval07+12} and evaluate it on \texttt{test2007}. We show the gap compared to the ImageNet supervised counterpart in brackets. The increases of at least 0.5 are in green. The results of random init, supervised, and MoCo v2 are from~\cite{Moco}. The results of BYOL~\cite{byol} and SimCLR~\cite{SimCLR} are from ~\cite{simsiam}. '*' means that results are implemented by us. Other methods are from their original papers. Our method outperforms the supervised counterpart by 5.5 points and the MoCo v2 baseline by 2.0 points. 
	200$^{\dag}$: the model is trained with 100 epochs but with symmetric loss~\cite{byol}, which makes one more network forward pass for each image; therefore, the number of network's forward pass is the same as that of MoCo v2~\cite{Mocov2} with 200 epochs.
	} 
	\vspace{-1em}
	\tablestyle{1.8pt}{1.05}
	\begin{tabular}{x{52}|x{25}|x{50}|x{50}x{50}c}
	    \hline
		pre-train &
		epoch &
		AP &
		AP$_\text{50}$ &
		AP$_\text{75}$ & \\ 
		\shline
		\randinit & - & \resrand{33.8}{} & \resrand{60.2}{} & \resrand{33.1}{} & \\
		supervised & 100 & \ressup{53.5}{} & \ressup{81.3}{} & \ressup{58.8}{} & \\
		\hline
		InstDis*~\cite{instdis} & 200 & \reshl{55.2}{+}{1.7} & \res{80.9}{-}{0.4} & \reshl{61.2}{+}{2.4} & \\
		BYOL~\cite{byol} & 200 & \reshl{55.3}{+}{1.8} & \res{81.4}{+}{0.1} & \reshl{61.1}{+}{2.3} & \\
		PIRL*~\cite{misra2020self} & 200 & \reshl{55.5}{+}{2.0} & \res{81.0}{-}{0.3} & \reshl{61.3}{+}{2.5} & \\
		SimCLR~\cite{SimCLR} & 200 & \reshl{55.5}{+}{2.0} & \reshl{81.8}{+}{0.5} & \reshl{61.4}{+}{2.6} & \\				
		SwAV~\cite{swav} & 800 & \reshl{56.1}{+}{2.6} & \reshl{82.6}{+}{1.3} & \reshl{62.7}{+}{3.9} & \\				
		BoWNet~\cite{BoWNet} & 200 & \reshl{55.8}{+}{2.3} & \res{81.3}{+}{0.0} & \reshl{61.1}{+}{2.3} & \\
		SimSiam~\cite{simsiam} & 200 & \reshl{57.0}{+}{3.5} & \reshl{82.4}{+}{1.1} & \reshl{63.7}{+}{4.9} & \\
		MoCo~\cite{Moco} & 200 & \reshl{55.9}{+}{2.4} & \res{81.5}{+}{0.2} & \reshl{62.6}{+}{3.8} & \\
		MoCo v2~\cite{Mocov2} & 200 & \reshl{57.0}{+}{3.5} & \reshl{82.4}{+}{1.1} & \reshl{63.6}{+}{4.8} & \\
		MoCo v2~\cite{Mocov2} & 800 & \reshl{57.4}{+}{3.9} & \reshl{82.5}{+}{1.2} & \reshl{64.0}{+}{5.2} & \\
		Infomin~\cite{infomin} & 200 & \reshl{57.6}{+}{4.1} & \reshl{82.6}{+}{1.3} & \reshl{64.3}{+}{5.5} & \\
		DetCo~\cite{detco} & 200 & \reshl{57.8}{+}{4.3} & \reshl{82.6}{+}{1.3} & \reshl{64.2}{+}{5.4} & \\
		InstLoc~\cite{instloc} & 200 & \reshl{57.9}{+}{4.4} & \reshl{82.9}{+}{1.6} & \reshl{64.9}{+}{6.1} & \\
		DenseCL~\cite{densecl} & 200 & \reshl{58.7}{+}{5.2} & \reshl{82.8}{+}{1.5} & \reshl{65.2}{+}{6.4} & \\
		PixPro~\cite{pixpro} & 200$^{\dag}$ & \reshl{58.8}{+}{5.3} & \reshl{83.0}{+}{1.7} & \reshl{\textbf{66.5}}{+}{7.7} & \\
		\hline
		DUPR (ours) & 200 & \reshl{\textbf{59.0}}{+}{5.5} & \reshl{\textbf{83.2}}{+}{1.9} & \reshl{66.0}{+}{7.2} & \\
		\hline
	\end{tabular}
	\vspace{.em}
	\label{tab:voc_detection}
	\vspace{-1.em}
\end{table}
\newcommand{\cres}[1]{#1}
\newcommand{\cresrand}[1]{#1}

\newcommand{\cgap}[2]{
	\fontsize{6pt}{1em}\selectfont{(${#1}${#2})}
}
\newcommand{\cgaphl}[2]{
	\fontsize{6pt}{1em}\selectfont{\textcolor{Highlight}{(${#1}$\textbf{#2})}}
}
\newcommand{\cgapll}[2]{
	\fontsize{6pt}{1em}\selectfont{\textcolor{red}{(${#1}$\textbf{#2})}}
}
\renewcommand{\randinit}{\tablestyle{1pt}{1} \begin{tabular}{z{21}y{26}} \multicolumn{2}{c}{random init.} \end{tabular}}

\begin{table*}[t]
	\small
	\caption{\textbf{Object detection and instance segmentation fine-tuned on COCO}. All the compared methods are pre-trained for 200 epochs on ImageNet. The results of random init, supervised, and MoCo are from~\cite{Moco}. We implement other methods in the same standard Mask R-CNN with R-50-FPN setting following MoCo~\cite{Moco}. \blank{Note in the original implementation of DenseCL~\cite{densecl}, InstLoc~\cite{instloc}, and PixPro~\cite{pixpro}, they evaluate results on a variant of Mask R-CNN with R-50-FPN by using a different RoI box head, which is different from the settings in MoCo~\cite{Moco}. So we download their pre-trained models and evaluate them in the same standard-setting. The increases of at least 0.5 points are in green while opposites are in red.}  
	}
	\vspace{-2.em}
	\hspace{-1.9em}
	\resizebox{1.05\linewidth}{!}{
		\subfloat[Mask R-CNN, R50-\textbf{FPN}, \textbf{1$\x$} schedule]{
			\tablestyle{0.8pt}{1.05}
			\begin{tabular}{cr|
					z{17}y{18}
					z{17}y{18}
					z{17}y{18}|
					z{17}y{18}
					z{17}y{18}
					z{17}y{18}c
				}
				\hline
				pre-train & ~ &
				\multicolumn{2}{c}{\apbbox{~}} &
				\multicolumn{2}{c}{\apbbox{50}} &
				\multicolumn{2}{c|}{\apbbox{75}} &
				\multicolumn{2}{c}{\apmask{~}} &
				\multicolumn{2}{c}{\apmask{50}} &
				\multicolumn{2}{c}{\apmask{75}} &\\
				\shline
				\demph{\randinit} & ~ &
				\demph{31.0} & ~ & \demph{49.5} & ~ & \demph{33.2} & ~ &
				\demph{28.5} & ~ & \demph{46.8} & ~ & \demph{30.4} & ~ & \\
				supervised & ~ &
				38.9 & ~ & 59.6 & ~ & 42.7 & ~ &
				35.4 & ~ & 56.5 & ~ & 38.1 & ~ & \\					
				\hline
				InstDis~\cite{instdis} & ~ &
				37.5 & \cgapll{-}{1.4} & 57.6 & \cgapll{-}{2.0} & 40.6 & \cgapll{-}{2.1} &
				34.1 & \cgapll{-}{1.3} & 54.7 & \cgapll{-}{1.8} & 36.5 & \cgapll{-}{1.6} \\
				
				PIRL~\cite{misra2020self} & ~ &
				37.6 & \cgapll{-}{1.3} & 57.7 & \cgapll{-}{1.9} & 41.1 & \cgapll{-}{1.6} &
				34.1 & \cgapll{-}{1.3} & 54.7 & \cgapll{-}{1.8} & 36.2 & \cgapll{-}{1.9} \\					
				SwAV~\cite{swav} & ~ &
				38.6 & \cgap{-}{0.3} & 60.5 & \cgaphl{+}{0.9} & 41.5 & \cgapll{-}{1.2} &
				35.5 & \cgap{+}{0.1} & 57.1 & \cgaphl{+}{0.6} & 37.8 & \cgap{-}{0.3} \\					
				MoCo~\cite{Moco} & ~ &
				38.5 & \cgap{-}{0.4} & 58.9 & \cgapll{-}{0.7} & 42.0 & \cgapll{-}{0.7} &
				35.1 & \cgap{-}{0.3} & 55.9 & \cgapll{-}{0.6} & 37.7 & \cgap{-}{0.4} \\
				
				MoCo v2~\cite{Mocov2} & ~ &
				38.9 & \cgap{+}{0.0} & 59.4 & \cgap{-}{0.2} & 42.4 & \cgap{-}{0.3} &
				35.5 & \cgap{+}{0.1} & 56.5 & \cgap{+}{0.0} & 38.2 & \cgap{+}{0.1} \\			
				
				VADeR~\cite{VADeR} & ~ &
				39.2 & \cgap{+}{0.3} & 59.7 & \cgap{+}{0.1} & 42.7 & \cgap{+}{0.0} &
				35.6 & \cgap{+}{0.2} & 56.7 & \cgap{+}{0.2} & 38.2 & \cgap{+}{0.1} \\	

				DenseCL~\cite{densecl} & ~ &
				39.4 & \cgaphl{+}{0.5} & 59.9 & \cgap{+}{0.3} & 42.7 & \cgap{+}{0.0} &
				35.6 & \cgap{+}{0.2} & 56.7 & \cgap{+}{0.2} & 38.2 & \cgap{+}{0.1} \\		

				InstLoc~\cite{instloc} & ~ &
				39.3 & \cgap{+}{0.4} & 59.8 & \cgap{+}{0.2} & 42.9 & \cgap{+}{0.2} &
				35.7 & \cgap{+}{0.3} & 56.9 & \cgap{+}{0.4} & 38.4 & \cgap{+}{0.3} \\

				DetCo~\cite{detco} & ~ &
				\textbf{40.1} & \cgaphl{+}{1.2} & \textbf{61.0} & \cgaphl{+}{1.4} & \textbf{43.9} & \cgaphl{+}{1.2} &
				\textbf{36.4} & \cgaphl{+}{1.0} & \textbf{58.0} & \cgaphl{+}{1.5} & \textbf{38.9} & \cgaphl{+}{0.8} \\

				PixPro~\cite{pixpro} & ~ &
				39.7 & \cgaphl{+}{0.8} & 60.0 & \cgap{+}{0.4} & 43.5 & \cgaphl{+}{0.8} &
				36.1 & \cgaphl{+}{0.7} & 57.1 & \cgap{+}{0.6} & \textbf{38.9} & \cgap{+}{0.8} \\
				
				\hline
				
				DUPR (ours) & ~ &
				40.0 & \cgaphl{+}{1.1} & 60.4 & \cgaphl{+}{0.8} & 43.4 & \cgaphl{+}{0.7} &
				36.2 & \cgaphl{+}{0.8} & 57.6 & \cgaphl{+}{1.1} & \textbf{38.9} & \cgaphl{+}{0.8} \\
				\hline
			\end{tabular}	
		}  
		
		\subfloat[Mask R-CNN, R50-\textbf{C4}, \textbf{1$\x$} schedule]{
			\tablestyle{.8pt}{1.05}
			\begin{tabular}{
					z{17}y{18}
					z{17}y{18}
					z{17}y{18}|
					z{17}y{18}
					z{17}y{18}
					z{17}y{18}c
				}
				\hline
				\multicolumn{2}{c}{\apbbox{~}} &
				\multicolumn{2}{c}{\apbbox{50}} &
				\multicolumn{2}{c|}{\apbbox{75}} &
				\multicolumn{2}{c}{\apmask{~}} &
				\multicolumn{2}{c}{\apmask{50}} &
				\multicolumn{2}{c}{\apmask{75}} &\\
				\shline
				\demph{26.4} & ~ & \demph{44.0} & ~ & \demph{27.8} & ~ &
				\demph{29.3} & ~ & \demph{46.9} & ~ & \demph{30.8} & ~ & \\					
				38.2 & ~ & 58.2 & ~ & 41.2 & ~ &
				33.3 & ~ & 54.7 & ~ & 35.2 & ~ & \\
				\hline
				37.8 & \cgap{-}{0.4} & 57.0 & \cgapll{-}{1.2} & 41.0 & \cgap{-}{0.2} &
				33.1 & \cgap{-}{0.2} & 54.2 & \cgapll{-}{0.5} & 35.3 & \cgap{+}{0.1} \\
				37.4 & \cgapll{-}{0.8} & 56.6 & \cgapll{-}{1.6} & 40.3 & \cgapll{-}{0.9} &
				32.8 & \cgapll{-}{0.5} & 53.4 & \cgapll{-}{1.3} & 34.8 & \cgap{-}{0.4} \\	
				33.0 & \cgapll{-}{5.2} & 54.3 & \cgapll{-}{3.9} & 34.6 & \cgapll{-}{6.6} &
				29.5 & \cgapll{-}{3.8} & 50.4 & \cgapll{-}{4.3} & 30.4 & \cgapll{-}{4.8} \\						
				38.5 & \cgap{+}{0.3} & 58.3 & \cgap{+}{0.1} & 41.6 & \cgap{+}{0.4} &
				33.6 & \cgap{+}{0.3} & 54.8 & \cgap{+}{0.1} & 35.6 & \cgap{+}{0.4} & \\
				
				38.9 & \cgaphl{+}{0.7} & 58.5 & \cgap{+}{0.3} & 42.1 & \cgaphl{+}{0.9} &
				34.2 & \cgaphl{+}{0.9} & 55.2 & \cgaphl{+}{0.5} & 36.6 & \cgaphl{+}{1.4} \\		
				
				%
				- &  & - &  & - &  &
				- &  & - &  & - &  \\
				
				39.3 & \cgaphl{+}{1.1} & 58.8 & \cgaphl{+}{0.6} & 42.5 & \cgaphl{+}{1.3} &
				34.3 & \cgaphl{+}{1.0} & 55.3 & \cgaphl{+}{0.6} & 36.7 & \cgaphl{+}{1.5} \\	
				
				39.5 & \cgaphl{+}{1.3} & 59.1 & \cgaphl{+}{0.9} & 42.7 & \cgaphl{+}{1.5} &
				34.5 & \cgaphl{+}{1.2} & 56.0 & \cgaphl{+}{1.3} & 36.8 & \cgaphl{+}{1.6} \\

				39.8 & \cgaphl{+}{1.0} & 59.7 & \cgaphl{+}{1.5} & 43.0 & \cgaphl{+}{1.8} &
				34.7 & \cgaphl{+}{1.4} & 56.3 & \cgaphl{+}{1.6} & 36.7 & \cgaphl{+}{1.5} \\

				40.0 & \cgaphl{+}{1.8} & 59.4 & \cgaphl{+}{1.2} & 43.2 & \cgaphl{+}{2.0} &
				34.8 & \cgaphl{+}{1.5} & 56.1 & \cgaphl{+}{1.4} & \textbf{37.3} & \cgaphl{+}{2.1} \\
				
				\hline
				\textbf{40.1} & \cgaphl{+}{1.9} & \textbf{59.8} & \cgaphl{+}{1.6} & \textbf{43.6} & \cgaphl{+}{2.4} &
				\textbf{34.9} & \cgaphl{+}{1.6} & \textbf{56.5} & \cgaphl{+}{1.8} & \textbf{37.3} & \cgaphl{+}{2.1} \\ \hline
			\end{tabular}	
		}  
	}  
	
	\vspace{-.5em}
	\hspace{-1.9em}
	\resizebox{1.05\linewidth}{!}{
		\subfloat[Mask R-CNN, R50-\textbf{FPN}, \textbf{2$\x$} schedule]{
			\tablestyle{.8pt}{1.05}
			\begin{tabular}{cr|
					z{17}y{18}
					z{17}y{18}
					z{17}y{18}|
					z{17}y{18}
					z{17}y{18}
					z{17}y{18}c
				}\hline
				pre-train & ~ &
				\multicolumn{2}{c}{\apbbox{~}} &
				\multicolumn{2}{c}{\apbbox{50}} &
				\multicolumn{2}{c|}{\apbbox{75}} &
				\multicolumn{2}{c}{\apmask{~}} &
				\multicolumn{2}{c}{\apmask{50}} &
				\multicolumn{2}{c}{\apmask{75}} &\\
				\shline
				\demph{\randinit} & ~ &
				\demph{36.7} & ~  & \demph{56.7} & ~  & \demph{40.0} & ~  &
				\demph{33.7} & ~  & \demph{53.8} & ~  & \demph{35.9} & ~  & \\
				supervised & ~ &
				40.6 & ~ & 61.3 & ~ & 44.4 & ~ &
				36.8 & ~ & 58.1 & ~ & 39.5 & ~ & \\
				MoCo v2~\cite{Mocov2} & ~ &
				40.9 & \cgap{+}{0.3} & 61.5 & \cgap{+}{0.2} & 44.7 & \cgap{+}{0.3} &
				37.0 & \cgap{+}{0.2} & 58.7 & \cgaphl{+}{0.6} & 39.8 & \cgap{+}{0.4} \\
				\hline
				DUPR (ours) & ~ &
				\textbf{41.7} & \cgaphl{+}{1.0} & \textbf{62.3} & \cgaphl{+}{1.0} & \textbf{45.2} & \cgaphl{+}{0.8} &
				\textbf{37.5} & \cgaphl{+}{0.7} & \textbf{59.2} & \cgaphl{+}{1.1} & \textbf{40.2} & \cgaphl{+}{0.7} & \\ \hline
				
			\end{tabular}	
		}  
		
		\subfloat[Mask R-CNN, R50-\textbf{C4}, \textbf{2$\x$} schedule]{
			\tablestyle{.8pt}{1.05}
			\begin{tabular}{
					z{17}y{18}
					z{17}y{18}
					z{17}y{18}|
					z{17}y{18}
					z{17}y{18}
					z{17}y{18}c
				}\hline
				\multicolumn{2}{c}{\apbbox{~}} &
				\multicolumn{2}{c}{\apbbox{50}} &
				\multicolumn{2}{c|}{\apbbox{75}} &
				\multicolumn{2}{c}{\apmask{~}} &
				\multicolumn{2}{c}{\apmask{50}} &
				\multicolumn{2}{c}{\apmask{75}} &\\
				\shline
				\demph{35.6} & ~  & \demph{54.6} & ~  & \demph{38.2} & ~  &
				\demph{31.4} & ~  & \demph{51.5} & ~  & \demph{33.5} & ~  & \\
				40.0 & ~ & 59.9 & ~ & 43.1 & ~ &
				34.7 & ~ & 56.5 & ~ & 36.9 & ~ & \\					
				41.0 & \cgaphl{+}{1.0} & 60.5 & \cgaphl{+}{0.6} & 44.5 & \cgaphl{+}{1.4} &
				35.7 & \cgaphl{+}{1.0} & 57.3 & \cgaphl{+}{0.8} & 38.1 & \cgaphl{+}{1.2} & \\
				\hline
				\textbf{41.5} & \cgaphl{+}{1.5} & \textbf{61.2} & \cgaphl{+}{1.3} & \textbf{45.1} & \cgaphl{+}{2.0} &
				\textbf{36.0} & \cgaphl{+}{1.3} & \textbf{58.0} & \cgaphl{+}{1.5} & \textbf{38.5} & \cgaphl{+}{1.6} & \\ \hline
			\end{tabular}	
		}  
	}  
	\label{tab:coco}
\end{table*}

\renewcommand{\randinit}{\tablestyle{1pt}{1} \begin{tabular}{z{21}y{26}} \multicolumn{2}{c}{random init.} \end{tabular}}

\definecolor{Gray}{gray}{0.5}

\definecolor{Highlight}{HTML}{39b54a}  

\renewcommand{\hl}[1]{\textcolor{Highlight}{#1}}

\begin{table}[t]
	\small
	\caption{\textbf{Object detection of RetinaNet fine-tuned on COCO}. All the unsupervised models are pre-trained on ImageNet for 200 epochs. All the results are implemented by us in the same settings.
	}
	\vspace{-2em}
	\centering
	\subfloat[RetinaNet, R50-\textbf{FPN}, \textbf{1$\x$} schedule]{
		\tablestyle{1pt}{1.0}
		\begin{tabular}{x{56}|x{54}|x{54}x{54}c}
		    \hline
			pre-train &
			AP &
			AP$_\text{50}$ &
			AP$_\text{75}$ & \\ 
			\shline
			\randinit & \resrand{24.5}{} & \resrand{39.0}{} & \resrand{25.7}{} & \\
			supervised  & \ressup{37.4}{} & \ressup{56.5}{} & \ressup{39.7}{} & \\
			MoCo v2~\cite{Mocov2} & \res{37.4}{+}{0.0} & \res{56.5}{+}{0.0} & \res{40.0}{+}{0.3} & \\
			\hline
			DUPR (ours) & \reshl{\textbf{38.1}}{+}{0.7} & \reshl{\textbf{57.3}}{+}{0.8} & \reshl{\textbf{41.1}}{+}{1.4} & \\ \hline
		\end{tabular}	
	} 
	\\
	\vspace{-.7em}
	\subfloat[RetinaNet, R50-\textbf{FPN}, \textbf{2$\x$} schedule]{
		\tablestyle{1pt}{1.0}
		\begin{tabular}{x{56}|x{54}|x{54}x{54}c}
		    \hline
			pre-train &
			AP &
			AP$_\text{50}$ &
			AP$_\text{75}$ & \\ 
			\shline
			\randinit & \resrand{32.2}{} & \resrand{49.4}{} & \resrand{34.2}{} & \\
			supervised & \ressup{38.9}{} & \ressup{58.5}{} & \ressup{41.5}{} & \\
			MoCo v2~\cite{Mocov2} & \reshl{39.4}{+}{0.5} & \reshl{59.0}{+}{0.5} & \reshl{42.2}{+}{0.7} & \\
			\hline
			DUPR (ours) & \reshl{\textbf{40.0}}{+}{1.1} & \reshl{\textbf{59.6}}{+}{1.1} & \reshl{\textbf{43.0}}{+}{1.5} & \\
			\hline
		\end{tabular}	
	} 
	\label{tab:retinanet}
\end{table}

\subsection{Pascal VOC Object Detection}\label{sec:vocdetection} 
\subsubsection{Experimental Setup}
PASCAL VOC~\cite{PASCALVOC} is a widely used small dataset for object detection, on which training from scratch can not catch up the performance compared to the pre-trained counterparts even with longer training~\cite{rethinking}. We fine-tune the Faster R-CNN with R-50-C4 backbone on Pascal VOC \texttt{trainval07+12} and evaluate the results on \texttt{test2007}. All the settings are the same as MoCo~\cite{Moco}. The RPN is built on conv\_4x feature map and R-CNN is built on the conv\_5x feature map in this detector. All the parameters of the network are fine-tuned end-to-end. The image size is [480, 800] in the training and 800 at inference. During fine-tuning, we train and synchronize all batch normalization layers. The batch normalization is used in the newly initialized RoI head layer. The fine-tuning takes a total of 24k iterations.

\subsubsection{Results Comparisons}

The results in Tab.~\ref{tab:voc_detection} show that DUPR outperforms other unsupervised methods and supervised counterparts. Most unsupervised methods outperform the supervised counterpart in AP$_\text{75}$ (\textit{which requires high localization accuracy}), which indicates that the representations learned by supervised classification may lose much information irrelevant to classification but useful for localization. However, the previous unsupervised pre-training methods are still designed for classification. In contrast, our DUPR is designed to encode the spatial information explicitly. DUPR pre-training outperforms the MoCo v2 strong baseline by 2.4 points in AP$_\text{75}$ and 0.8 points in AP$_\text{50}$, and further significantly improves the localization accuracy. DUPR also obtains state-of-the-art performance in AP and AP$_\text{50}$. When compared to the pre-training by supervised classification, DUPR largely improves the AP$_\text{75}$ by 7.3 points. It verifies that DUPR contains more spatial information than MoCo v2 and ImageNet supervised pre-training.

\subsection{COCO Object Detection and Segmentation}\label{sec:coco_results}
\subsubsection{Experimental Setup}
We compare the fine-tuning results of Mask R-CNN R-50-FPN, Mask R-CNN R-50-C4, and RetinaNet R-50-FPN with other unsupervised and ImageNet supervised counterparts, including both one-stage and two-stage detectors with different backbones. We fine-tune these detectors on COCO \texttt{train2017} with 118k images, and test on COCO \texttt{val2017}. 
For all detectors, input images are randomly resized to a scale within [640, 800] during training and fixed at 800 for inference. All the layers are trained end-to-end. 
For Mask R-CNN R-50-FPN and Mask R-CNN R-50-C4, we strictly follow the settings in MoCo~\cite{Moco}. For Mask R-CNN R-50-FPN and RetinaNet R-50-FPN, the batch normalization is used in the newly initialized FPN.
Other parameters of RetinaNet follow the default setting of Detectron2~\cite{detectron2}. We explore the standard 1$\x$ and 2$\x$ schedule for these detectors.
For Mask R-CNN R-50-FPN, we also compare the fine-tuned results with the strong baseline MoCo v2~\cite{Mocov2} at fewer training iterations (12k, 18k, and 36k iterations) to study the convergence speed. 

\subsubsection{Mask R-CNN R-50-FPN} The results of Mask R-CNN, R-50-FPN, 1$\x$ schedule are shown in Tab.~\ref{tab:coco} (a). DUPR outperforms other unsupervised methods and the supervised counterparts (\eg,~surpasses MoCo v2 baseline by 1.1 points in mAP). In 2$\x$ schedule, DUPR outperforms MoCo v2 by 0.7 points in mAP and the supervised counterpart by 1.0 point in mAP as shown in Tab.~\ref{tab:coco} (c).

\subsubsection{Mask R-CNN R-50-C4}	
As shown in Tab.~\ref{tab:coco} (b), in 1$\x$ schedule, DUPR outperforms all other unsupervised and supervised counterparts (\eg,~surpasses MoCo v2 by 1.2 points in mAP). Compared to the ImageNet supervised pre-training, the gain in AP$_\text{75}$ is larger than AP$_\text{50}$ (2.4 points v.s. 1.9 points), indicating that DUPR pre-training improves the localization ability. 
In 2$\x$ schedule, where pre-training is less important, our method still outperforms the MoCo v2 by 0.5 points and the ImageNet supervised pre-training by 1.5 points in mAP as shown in Tab.~\ref{tab:coco} (d).

\subsubsection{RetinaNet R-50-FPN}
We choose to fine-tune the RetinaNet~\cite{lin2017focal} with R-50-FPN on COCO in 1$\x$ and 2$\x$ schedule.	As shown in Tab.~\ref{tab:retinanet}, in the 1$\x$ schedule, MoCo v2 has the same AP as the supervised counterpart. Our method outperforms MoCo v2 pre-training and supervised counterpart by 0.7 points in AP. Our method generalizes well on the one-stage object detector.

\subsubsection{Fine-tune with Fewer Iterations}
Pre-training can speed up the convergence of the object detectors~\cite{rethinking}. So, we explore the performance of different unsupervised pre-trained models, when serving as initialization for fine-tuning Mask R-CNN R-50-FPN at iterations of 12k, 18k, 36k, 90k, and 180k in Fig.~\ref{fig:iteration_accuracy}. Our DUPR outperforms MoCo v2 and even the ImageNet supervised pre-training at all iterations. When fine-tuning with only 12k iterations, DUPR significantly outperforms the MoCo v2 by 2.9 points in mAP. It indicates that DUPR provides a better initialization and faster convergence speed than other methods. When fine-tuning with 90 iterations, DUPR still outperforms the supervised counterpart by 1.1 points.

\subsection{Other Localization-Sensitive Tasks}
\label{sec:othertask}
\begin{table}[t]
	\caption{\textbf{Comparison with other pre-training methods, fine-tuned on various localization-sensitive tasks.} 
	Our method outperforms MoCo v2 baseline and supervised counterparts. The seg. denotes the semantic segmentation.} 
	\vspace{-1.3em}
	\tablestyle{1.8pt}{1.05}
	\begin{tabular}{ry{24}|x{40}x{50}|x{70} c}
	    \hline
		\multicolumn{2}{c|}{} &
		\multicolumn{2}{c|}{\fontsize{7.5pt}{1em}\selectfont \textbf{Cityscapes Instance seg.}} & 
		\fontsize{7.5pt}{1em}\selectfont \textbf{Cityscapes Seg.}
		\\ \hline
		\multicolumn{2}{c|}{pre-train} &
		\apmask{~} & \apmask{50} &  
		mIoU
		\\
		\shline
		\multicolumn{2}{c|}{\demph{random init.}}
		& \resrand{25.4}{} & \resrand{51.1}{}  
		& \resrand{65.3}{}  
		& \\
		\multicolumn{2}{c|}{supervised}
		& \ressup{32.9}{} & \ressup{59.6}{}  
		& \ressup{74.6}{}  
		& \\
		\multicolumn{2}{c|}{MoCo v2~\cite{Mocov2}}
		& \reshl{33.9}{+}{1.0} & \reshl{60.8}{+}{1.2}  
		& \reshl{75.7}{+}{1.1}  
		& \\
		\hline
		\multicolumn{2}{c|}{DUPR (ours)}
		& \reshl{\textbf{34.4}}{+}{1.5} & \reshl{62.3}{+}{2.7}  
		& \reshl{\textbf{76.7}}{+}{2.1}  
		& \\
		\hline
	\end{tabular}
	\label{tab:cityscape}
	\vspace{1.3em}
	\begin{tabular}{ry{28}|x{48}x{48}x{48} c}
	\hline
    \multicolumn{2}{c|}{} &
    \multicolumn{3}{c}{\fontsize{7.5pt}{1em}\selectfont \textbf{COCO Keypoint Detection}} & 
    \\ \hline
    \multicolumn{2}{c|}{pre-train} &
    \apkp{~} & \apkp{50} & \apkp{75} &  
    \\
    \shline
    \multicolumn{2}{c|}{\demph{random init.}}
    & \resrand{65.9}{} & \resrand{86.5}{} & \resrand{71.7}{}  
    & \\
    \multicolumn{2}{c|}{supervised}
    & \ressup{65.8}{} & \ressup{86.9}{} & \ressup{71.9}{}  
    & \\
    \multicolumn{2}{c|}{MoCo v2}
    & \reshl{66.8}{+}{1.0} & \res{87.3}{+}{0.4} & \reshl{73.1}{+}{1.2}  
    & \\
    \hline
    DUPR & IN-1M
    & \reshl{\textbf{67.1}}{+}{1.3} & \reshl{\textbf{87.4}}{+}{0.5} & \reshl{\textbf{72.9}}{+}{1.0}  
    & \\ \hline
    \end{tabular}
	\label{tab:kp}

    \vspace{1.3em}
	\begin{tabular}{ry{28}|x{48}x{48}x{48} c}
    \hline
    \multicolumn{2}{c|}{} &
    \multicolumn{3}{c}{\fontsize{7.5pt}{1em}\selectfont
    \textbf{LVIS-v1.0 Instance Seg.}} & 
    \\ \hline
    \multicolumn{2}{c|}{pre-train} &
    			\apmask{~} &
			\apmask{50} &
			\apmask{75} & 
    \\
    \shline
    \multicolumn{2}{c|}{\demph{random init.}}
    & \resrand{19.1}{} & \resrand{29.8}{} & \resrand{20.2}{}  
    & \\
    \multicolumn{2}{c|}{supervised}
    & \ressup{22.3}{} & \ressup{34.7}{} & \ressup{23.5}{}  
    & \\
    \multicolumn{2}{c|}{MoCo v2}
    & \reshl{22.8}{+}{0.5} & \reshl{35.2}{+}{0.5} & \reshl{24.4}{+}{0.9}  
    & \\
    \hline
    \multicolumn{2}{c|}{DUPR}
    & \reshl{\textbf{23.8}}{+}{1.5} & \reshl{\textbf{36.1}}{+}{1.4} & \reshl{\textbf{25.3}}{+}{1.8}  
    & \\ \hline
    \end{tabular}
	\label{tab:lvis}
\end{table}
\subsubsection{Instance Seg. on Cityscapes} Cityscapes~\cite{Cityscapes} is a dataset that focuses on semantic understanding of urban street scenes. We fine-tune the Mask R-CNN R-50-FPN following the settings in MoCo~\cite{Moco}. Batch normalization is added before the FPN. All layers are trained end-to-end. Other hyperparameters follow the default settings of Detectron2~\cite{detectron2}. We fine-tune the model on the \texttt{train\_fine} set (2975 images) for 90k iterations, and test on the \texttt{val} set. DUPR outperforms MoCo v2 by 0.5 points in mAP as shown in Tab.~\ref{tab:cityscape}, which indicates the DUPR pre-training has a good generalization.

\subsubsection{Semantic Seg. on Cityscapes}
For the semantic segmentation on Cityscapes, we follow the FCN-based structure and settings used in MoCo~\cite{Moco}. 
The FCN-based structure consists of the convolutional layers in Resnet 50 and the $3\times 3$ convolutions with dilation 2 and stride 1 in conv5\_x stage. Then it is followed by two extra $3\times 3$ convolutions of 256 channels with dilation 6, and a $1\times 1$ convolution is added for per-pixel classification. 
Since there is no officially released code, we reimplement the FCN-based structure. DUPR outperforms MoCo v2 by 0.9 points as shown in Tab.~\ref{tab:cityscape}. The results on Cityscapes indicate that DUPR pre-training transfers well to other \textit{region-level} tasks.

\subsubsection{Keypoint Detection on COCO}
The task of keypoint detection is to simultaneously detect people and locate their keypoints. We fine-tune the Mask R-CNN R-50-FPN (keypoint version) on COCO \texttt{train2017} and evaluate it on COCO \texttt{val2017}, following~\cite{Moco}. DUPR outperforms the supervised counterpart by 1.3 points in mAP as shown in Tab.~\ref{tab:kp}. 

\subsubsection{Instance Seg. on LVIS-v1.0}\label{sec:lvis} LVIS~\cite{lvis} is an instance segmentation dataset, which has 1203 long tail distributed categories and provides high-quality segmentation masks. 
We fine-tune the Mask R-CNN R-50-FPN on LVIS \texttt{train\_v1.0} and test on LVIS \texttt{val\_v1.0}. We train the model for a total of 180k iterations. DUPR outperforms MoCo v2 by 1.0 point in \apmask{~} as shown in Tab.~\ref{tab:lvis}.
\subsection{Object Detection v.s. Classification} \label{sec:cls_acc}
Object detection includes both classification and localization. To better understand why DUPR improves object detection, we also report the results of linear probing on ImageNet for reference to obtain more insight. We compare the classification accuracy and detection mAP for various unsupervised pre-training methods in Tab.~\ref{tab:detvscls}. \revision{We notice that under these settings, with the compared models, there exists little correlation between ImageNet classification and object detection performance.}
For example, MoCo v2~\cite{Mocov2} is lower than BYOL~\cite{byol} by 3.1 points in ImageNet accuracy but higher than BYOL~\cite{byol} by 1.7 points in VOC AP.
DUPR and DenseCL~\cite{densecl} have a drop of 3.7 points in ImageNet linear evaluation compared to the MoCo v2 baseline. This drop is likely resulted from the joint optimization of both \textit{patch-level} and \textit{image-level} contrastive loss, which is more difficult and will affect the optimization of \textit{image-level} contrastive loss. A better balance between global representations for classification and local representations for localization is possible but is not the focus of this paper.
We can conclude that the improvements of DUPR in object detection are not from the ability of better classification but the ability of better localization.

\begin{table}[t]
	\caption{\textbf{Object detection v.s. Classification.} The \textit{ImageNet cls.} means linear evaluation on ImageNet. For COCO and VOC detection, we report AP of Mask R-CNN with R-50-C4 and Faster R-CNN with R-50-C4, respectively. All the self-supervised models are pre-trained with 200 epochs. The results of SimCLR~\cite{SimCLR}, BYOL~\cite{byol}, SwAV~\cite{swav}, and SimSiam~\cite{simsiam} are from ~\cite{simsiam}. Note that the DenseCL~\cite{densecl} is a concurrent self-supervised learning method designed for object detection. We can see from this table that \revision{the improvements in the ImageNet linear evaluation do not always lead to the improvements in the transfer performance to object detection.}  
	}
	\vspace{-1em}
	\tablestyle{1.8pt}{1.25}
	\centering
	\begin{tabular}{x{50}|x{50}|x{50}|x{50}c}
	    \hline
		pre-train &
		ImageNet cls. &
		COCO Det. &
		VOC Det. & \\ 
		\shline
		SimCLR~\cite{SimCLR} & 68.3 & 37.9 & 55.5 & \\
		BYOL~\cite{byol} & \textbf{70.6} & 37.9 & 55.3 & \\
		SwAV~\cite{swav} & 69.1 & 37.6 & 55.4 & \\
		MoCo v2~\cite{Mocov2} & 67.5 & 38.9 & 57.0 & \\ 
		SimSiam~\cite{simsiam} & 70.0 & 39.2 & 57.0 & \\ 
		\hline
		DenseCL~\cite{densecl} & 63.6 & 39.3 & 58.7 & \\ 
		DUPR & 63.6 & \textbf{40.1} & \textbf{59.0} & \\ \hline
	\end{tabular}
	\label{tab:detvscls}
\end{table}

\subsection{Ablation Experiments}\label{sec:experiments}	
\subsubsection{Experimental Setup}
The ablation experiments are conducted on PASCAL VOC with Faster R-CNN R-50-C4 and COCO with R-50-FPN. We also report the SVM classification on PASCAL VOC, following the settings in~\cite{goyal2019scaling}, where the feature is fixed and used to train an SVM classifier on PASCAL VOC classification task.

\subsubsection{Influence of $\alpha_{0:3}$ for Image Contrastive Loss}
\revision{In this ablation study, the $\beta_{0:3}$ is set as (0, 0, 0, 0). The results in Tab.~\ref{tab:deepsupervision_voc}~(a) show: (1) all configurations of $\alpha_m$ can improve the AP in VOC detection, especially for the high IoU metric AP$_\text{75}$, suggesting that intermediate level contrastive loss can improve the localization ability; (2) only the configuration of (0.1, 0.4, 0.7, 1.0) has improvements on both VOC and COCO detection. The reasons why large weights (\ie, configuration of (1, 1, 1, 1)) on shallow layers decrease the performance on COCO detection are two aspects: (1) large weights on shallow layers will influence the optimization of deep layers, which is more important for classification than shallow layers; (2) COCO contains more classes than VOC, and the performance on COCO relies more on the classification ability. In fact, setting the deep layers with larger weights is a common practice in~\cite{pspnet,OCNet,googlenet,mmseg2020}, although their tasks are different from ours.}

Now we study the classification ability of intermediate layers. It can be seen in Tab.~\ref{tab:deepsupervision_voc} (b) that all three configurations of $\alpha_{0:3}$ largely improve the classification performance of shallow layers. For example, when set $\alpha_{0:3}=(0.1,0.4,0.7,1.0)$, it improves the classification performance of conv3\_x by 10 points and conv4\_x by 6.3 points. We also notice a slight decrease in the classification performance of the conv5\_x, which makes sense as the optimization of multi-level contrastive loss is more challenging than single-level one.
Combining Tab.~\ref{tab:deepsupervision_voc}~(a) and Tab.~\ref{tab:deepsupervision_voc}~(b) we can conclude: better representations of shallow layers can improve transfer performance, especially for the localization aspect of object detection.


\begin{table}[t]
	\small
	\caption{\textbf{Ablation of $\mathbf{\mathcal{\alpha}_m}$ for image contrastive loss}. All models are pre-trained with 200 epochs on ImageNet. $\mathcal{\alpha}_m$ is the weight to balance each layer in Eq.~\eqref{eq:loss_all}. (a) Object detection is fine-tuned on Pascal VOC \texttt{trainval07+12} and tested on the Pascal VOC test 2007. (b) The SVM classification is fine-tuned on Pascal VOC 2007. The results show that there exists a correlation between the classification ability for intermediate layers and transferring ability to object detection.
	}
	\vspace{-2.em}
	\centering
	\subfloat[Faster R-CNN, R50-\textbf{C4}]{
		\tablestyle{1pt}{1.0}

		\begin{tabular}{x{45}|z{17}y{18}
		                      z{17}y{18}
		                      z{17}y{18}|
		                      z{17}y{18}
		                      z{17}y{18}
		                      z{17}y{18}}
		    \hline
			& \multicolumn{6}{c|}{VOC}
			& \multicolumn{4}{c}{COCO}
			\\ \hline
			$\alpha_{0:3}$ &
			\multicolumn{2}{c}{AP} &
			\multicolumn{2}{c}{AP$_\text{50}$} &
			\multicolumn{2}{c|}{AP$_\text{75}$} &
			\multicolumn{2}{c}{\apbbox{~}} &
            \multicolumn{2}{c}{\apmask{~}} \\
			\shline
			0,0,0,1  & 57.0 & ~ & 82.4 & ~ & 63.6 & ~ & 38.9 & ~ & 35.5 \\ \hline 
			0,0,1,1  & 57.4 & \cgap{+}{0.4} & 82.7 & \cgap{+}{0.3} & 64.1 & \cgaphl{+}{0.5} & 38.9 & \cgap{+}{0.0} & 35.0 & \cgapll{-}{0.5}  \\
			1,1,1,1  & 57.4 & \cgap{+}{0.4} & 82.3 & \cgap{-}{0.1} & 63.9 &\cgap{+}{0.3} & 38.5 &\cgap{-}{0.4} & 35.0 & \cgapll{-}{0.5} \\
			0.1,0.4,0.7,1 & \textbf{57.6} & \cgaphl{+}{0.6} & 82.4 & \cgap{+}{0.0} & \textbf{64.2} & \cgaphl{+}{0.6} & \textbf{39.2} & \cgap{+}{0.3} & \textbf{35.5} & \cgap{+}{0.0} \\
			\hline
		\end{tabular}

	} 
	\\
	\vspace{-.5em}
	\subfloat[SVM Classification of Different Levels]{
		\tablestyle{1pt}{1.0}
		\begin{tabular}{x{42}|x{45}|x{50}|x{45}|x{43}c}
		    \hline
			$\alpha_{0:3}$ &
			conv2\_x &
			conv3\_x & 
			conv4\_x &
			conv5\_x \\ 
			\shline
			0,0,0,1 & \ressup{47.3}{}{} & \ressup{58.8}{}{} & \ressup{74.0}{}{} & \ressup{\textbf{84.1}}{}{} & \\
			\hline
			0,0,1,1 & \res{47.6}{+}{0.3}  & \reshl{60.0}{+}{1.2} & \reshl{\textbf{81.0}}{+}{7.0} & \res{83.7}{-}{0.4} &\\
			1,1,1,1 & \res{\textbf{57.2}}{+}{9.9} & \reshl{\textbf{70.5}}{+}{11.7} & \reshl{80.3}{+}{6.3} & \resll{82.9}{-}{1.2} &\\
			0.1,0.4,0.7,1 & \reshl{52.1}{+}{4.8} & \reshl{68.8}{+}{10.0} & \reshl{80.9}{+}{6.9} & \res{83.7}{-}{0.4} & \\ \hline
		\end{tabular}
		
	} 
	\label{tab:deepsupervision_voc}
	\vspace{-1.em}
\end{table}

\subsubsection{Influence of RoI Size} 
\revision{In this ablation study, we set $\alpha_{0:3}=(0.1, 0.4, 0.7, 1.0)$ and $\beta_{0:3}=(0, 0, 0, 1)$.} \revision{The results of different RoI sizes are presented in Tab.~\ref{tab:patch_reidentification}~(a)}. 
We can see that $\beta_{0:3}=(0, 0, 0, 1)$ with RoI size of 1 does not have much difference compared to $\beta_{0:3}=(0, 0, 0, 0)$ in object detection, where the patch Re-ID degenerates to global view contrastive learning and has no constraint on the local representations, although it is spatially aligned. 
Large RoI size improves the performance, which indicates that larger RoI size can get more discriminative \textit{region-level} representations for object detection. 
\revision{ But it is not always the larger, the better: if the RoI size is \textit{larger than the size of a feature map}, it will not acquire more useful information. For example, the improvements become saturated when the RoI size is larger than 7, which is the size of conv5\_x feature map. So we set RoI size as 7 on conv5\_x by default. Similarly, the size of conv4\_x is 14, so we set the RoI size as 14 on conv4\_x by default.}

\subsubsection{Influence of $\beta_{0:3}$ for Patch Contrastive Loss}
\revision{In this ablation study, we set $\alpha_{0:3}=(0.1, 0.4, 0.7, 1.0)$.} Simply adding patch loss to conv5\_x (by setting $\beta_{0:3}=(0,0,0,1)$) improves the AP by 0.7 points in VOC and 0.3 points in COCO. And adding patch loss to both conv4\_x and conv5\_x \revision{(by setting $\beta_{0:3}=(0,0,1,1)$)} feature maps further improves the AP by 0.5 points in COCO and 0.2 points in VOC as shown in Tab.~\ref{tab:patch_reidentification} (b), which indicates that intermediate supervision and patch contrastive loss are complementary. 

\subsubsection{Image-level v.s. Patch-level Contrastive Loss}
We compare the single \textit{image-level} and \textit{patch-level} contrastive loss on conv5\_x in Tab.~\ref{tab:patch_reidentification} (c). \textit{Patch-level} contrastive loss improves the localization ability but slightly reduces the classification ability, compared to the \textit{image-level} contrastive loss. For example, in VOC detection, \textit{patch-level} contrastive loss significantly improves the AP$_\text{75}$ (which is more related to localization ability) by 1.0 point but slightly reduces the AP$_\text{50}$ by 0.2. It also slightly reduces the COCO AP by 0.1 points, since classification ability is more important to COCO than VOC (\textit{note COCO has 80 categories while VOC has 20 categories}). When we combine both \textit{image-level} and \textit{patch-level} contrastive loss and achieve a balance between classification and localization as shown in the bottom line of Tab.~\ref{tab:patch_reidentification} (b), we achieve the best performance.
\begin{table}[t]
	\small
	\caption{(a) \textbf{Ablation of RoI size.} A larger size of RoI has better results than a smaller one. We add patch loss to the conv5\_x feature map to do the RoI size ablation study on Pascal VOC, Faster R-CNN, R-50-C4 and on COCO, Mask R-CNN, R-50-FPN. (b) \textbf{Ablation of $\mathbf{\beta_{m}}$.} We use RoI sizes of 7 and 14 for conv5\_x and conv4\_x feature maps, respectively. (c) We compare single image-level and patch-level contrastive loss on conv5\_x.
}
	\vspace{-2em}
	\centering
	\subfloat[Ablation of RoI Size]{
		\tablestyle{1pt}{1.0}
		\begin{tabular}{x{40}|x{30}|x{30}|x{35}|x{35}x{35}c}
		    \hline
			& \multicolumn{3}{c|}{VOC}
			& \multicolumn{2}{c}{COCO} \\ \hline
			RoI Size &
			AP &
			AP$_\text{50}$ &
			AP$_\text{75}$ & 
			\apbbox{~} &
			\apmask{~}	\\ 
			\shline
							 1 & 57.7 & 82.4 & 63.9 & 39.1 & 35.4\\
			 3 & 58.1 & 82.5 & 65.0 & 39.3 & 35.6 \\
			 7 & 58.3 & 82.7 & 64.9 & \textbf{39.5} & \textbf{35.7} \\	
			 9 & \textbf{58.4} & 82.8 & \textbf{65.3} & 39.4 & \textbf{35.7} \\
			  11 & \textbf{58.4} & \textbf{82.9} & 65.2 & \textbf{39.5} & \textbf{35.7} \\
			\hline
		\end{tabular}	
	} 
	\\
	\vspace{-.8em}
	\subfloat[Ablation of $\beta_{0:3}$]{
		\tablestyle{1pt}{1.0}
		\begin{tabular}{x{45}|x{35}x{35}x{35}|x{35}x{35}} \hline
			& \multicolumn{3}{c|}{VOC}
			& \multicolumn{2}{c}{COCO}
			\\ \hline
			$\beta_{0:3}$ &
			AP &
			AP$_\text{50}$ & 
			AP$_\text{75}$ &
			\apbbox{~} &
			\apmask{~}  \\
			\shline
			 0,0,0,0 & 57.6 & 82.4 & 64.2 & 39.2 & 35.5 \\ \hline
			 0,0,0,1 & 58.3 & 82.7 &  64.9 & 39.5 & 35.7  \\
			 0,0,1,1 & \textbf{58.5} & \textbf{83.2} & \textbf{65.2} & \textbf{40.0} & \textbf{36.2} \\
			 1,1,1,1 & \textbf{58.5} & 82.8 & 65.2 & 39.7 & 36.0 \\ 
			 0.1,0.4,0.7,1 & 58.3 & \textbf{83.5} & \textbf{65.3} & 39.9 & 36.1 \\ 
			 \hline
		\end{tabular}	
	} 
	\\
	\vspace{-.8em}
	\subfloat[Image and patch contrastive loss at conv5\_x]{
		\tablestyle{1pt}{1.0}
		\begin{tabular}{x{45}|x{35}x{35}x{35}|x{35}x{35}} \hline
			& \multicolumn{3}{c|}{VOC}
			& \multicolumn{2}{c}{COCO}
			\\ \hline
            &
			AP &
			AP$_\text{50}$ & 
			AP$_\text{75}$ &
			\apbbox{~} &
			\apmask{~}  \\
			\shline
			image-level & 57.0 & 82.4 & 63.6 & \textbf{38.9} & \textbf{35.5} \\       
			patch-level & \textbf{57.9} & 82.2 &  64.6 & 38.8 &  35.3 \\ \hline
		\end{tabular}	
	} 
	\label{tab:patch_reidentification}
	\vspace{-.3em}
\end{table}

\subsection{Visualization}\label{sec:visualization}
\subsubsection{Visualization of Spatial Sensitivity}
\begin{figure}[t!]
	\centering
	\includegraphics[width=.87\linewidth]{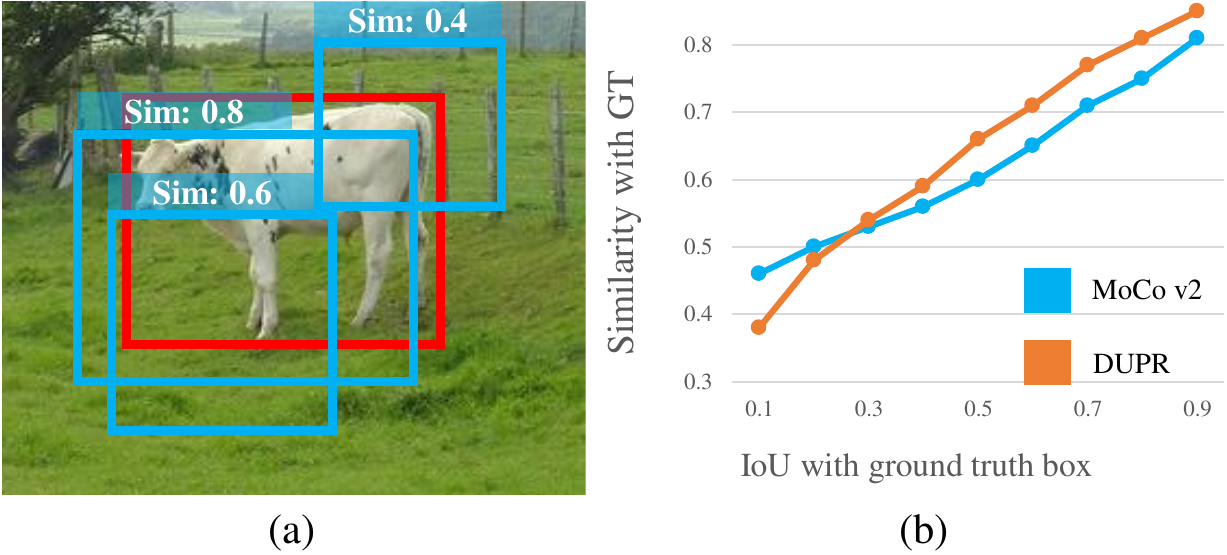}
	\vspace{-1mm}
	\caption{\textbf{The visualization of spatial sensitivity}. (a) We visualize the similarity between ground truth (Red bounding box) and RoIs (blue boxes with different IoUs) using the Eq.~\eqref{eq:sim}. (b) shows that DUPR learns more spatial-sensitive features than MoCo v2: the curve of DUPR has a larger slope than that of MoCo v2. More specifically, DUPR has a higher similarity between RoI and ground truth box when their IoU is larger than 0.5, and has a smaller similarity than MoCo v2 when IoU is less than 0.3. It means that features learned by DUPR are more discriminative than MoCo v2 to suppress wrong predictions (boxes with small IoU).
		\label{fig:iou_similarity}}
	\vspace{-1.em}
\end{figure}
\begin{figure*}[t]\centering
	\includegraphics[width=1.0\linewidth]{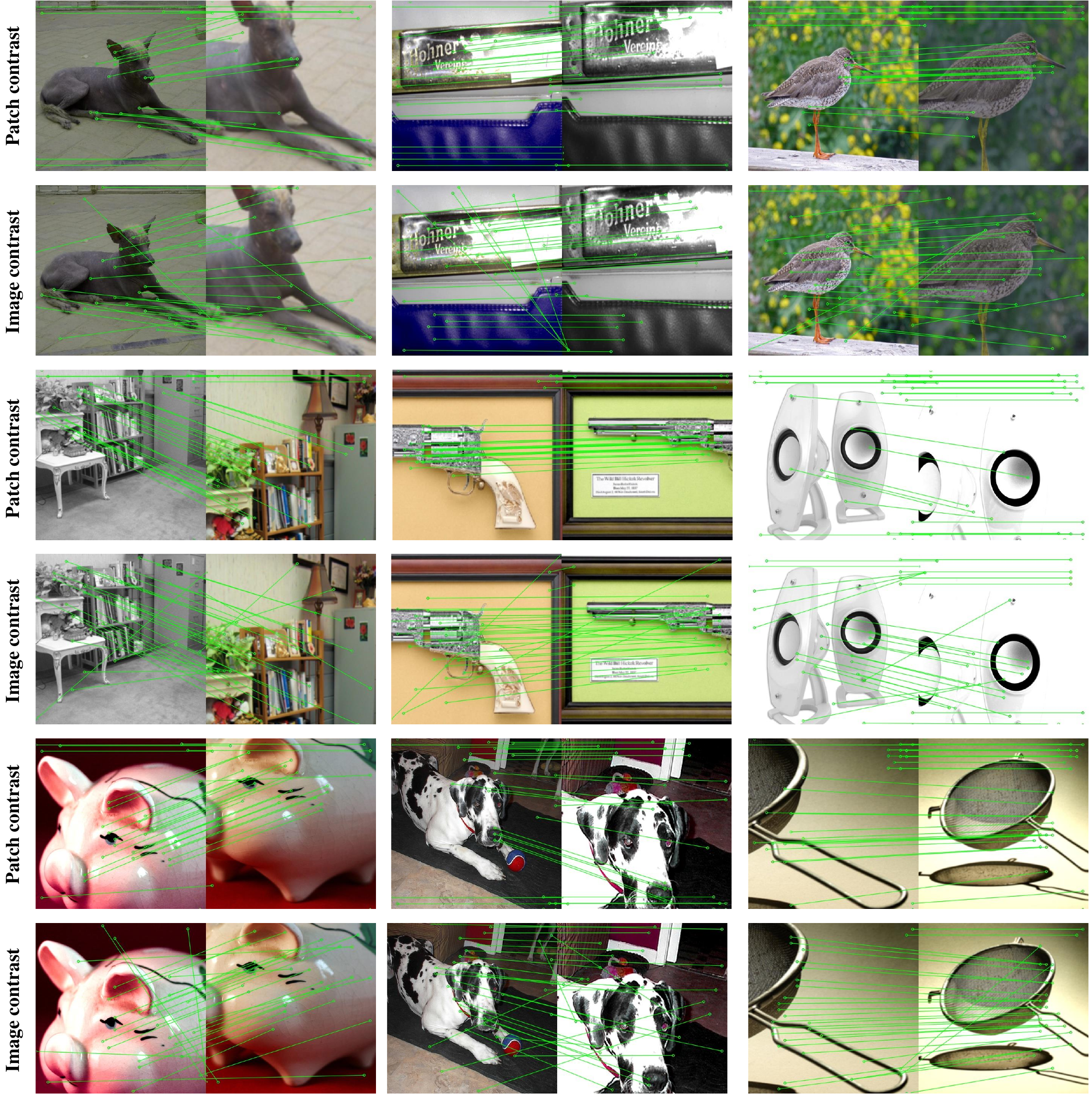}
	\vspace{-.8em}
	\caption{\textbf{Visualization of Correspondence.} \revision{We use pre-trained representations to match patches in two views, and compare the results of two self-supervised tasks: (1) \textit{patch-level} contrastive learning (patch Re-ID); (2) \textit{image-level} contrastive learning (MoCov2). For fair comparison, both two tasks only add loss on the conv5\_x with the same loss weight.} 
	For visualization, we use points to represent patches. It shows that representations of patch Re-ID can match the local patches in the other view better than MoCo v2.
	}
	\label{fig:keypoint}
	\vspace{-1.em}
\end{figure*}
To verify that patch Re-ID can learn spatial-sensitive features, we draw the IoU-similarity curve: each point in the curve is obtained by calculating the IoU and similarity between a RoI and the ground truth box. We randomly select 1000 images from the ImageNet validation set. For each image, we sample 20 RoIs with different IoUs (range from 0 to 1) between the ground truth box and RoIs. We then use the RoI Align to extract region features from conv4\_x feature map, according to the ground truth box and RoIs. Denote the feature extracted from a ground truth box as $q$ and a feature extracted from a RoI as $k_{\alpha}$, where the $\alpha$ denotes the IoU between the ground truth box and the RoI. Both $q$ and $k_{\alpha}$ are of shape $(C,S\times S)$. We calculate an average of cosine similarity between $q$ and $k_{\alpha}$ as:
\begin{equation}
	\small
	Sim(q, k_{\alpha}) = \frac{1}{S\times S} \sum_{p=1}^{S\times S} q_{p}{\cdot}k_{\alpha,p},
	\label{eq:sim}
\end{equation}
where $q_{p}$ and $k_{\alpha,p}$ are normalized features at position $p$ of region features. We compare DUPR with MoCo v2 in Fig.~\ref{fig:iou_similarity}. This visualization shows that DUPR is more sensitive to IoU: DUPR has a steeper slope than MoCo v2. When IoU between RoI and ground truth box is above 0.5 (usually assigned as positive samples in object detection~\cite{ren2015faster}), DUPR has a higher similarity; when IoU is below 0.3 (usually assigned as negative samples), DUPR has a lower similarity. Such property makes the representations easy to suppress the false-negative RoIs.

\subsubsection{Visualization of Correspondence}

We use the learned representations to visualize the correspondence between two views. First, we create two views via random data augmentation following the settings in ~\cite{Moco}. We set the image size to 448. We use the feature map of conv4\_x (which is of shape $(1024, 28, 28)$) for matching. For each patch (one of $28\times 28$) feature in one view, we find the patch with the highest similarity in the other view. For visualization, we use the point to represent a patch. The results shown in Fig.~\ref{fig:keypoint} demonstrate that patch Re-ID matches the corresponding patches better than MoCo v2.

\section{Conclusion}
This paper presents an unsupervised visual representation learning method, named DUPR, to bridge the gap between the unsupervised pre-training and downstream object detection tasks. Different from the previous methods that only learn discriminative \textit{image-level} representations in the final layer, our method learns discriminative \textit{region-level} multi-level representations. Therefore, our method outperforms other unsupervised models and even the supervised counterpart when transferred to downstream tasks related to object detection. Moreover, our method is robust to various object detectors, fine-tuning iterations. 
We hope our simple yet effective method could serve as a baseline of unsupervised pre-training for object detection tasks.

\revision{It is worth noticing that the DUPR presented in this paper learns representations for object detection from static images. While, rather than static images, we humans learn more from videos, which contain not only spatial configurations but also rich temporal information. Thus, it is of great interest to extend the proposed DUPR for learning spatial-temporal representations from videos in the future. Moreover, our current implementation of DUPR is based on MoCov2~\cite{Mocov2} and needs multiple memory banks. To save the memory occupation, we will further study how to apply DUPR to other self-supervised learning methods (\eg, BYOL~\cite{byol} or SimSiam~\cite{simsiam}). Beyond the pretext task, how to design more suitable data augmentation pipeline for dense prediction tasks can also be studied in the future.}

	{\small
		\bibliographystyle{IEEEtran}
		\bibliography{egbib}
	}

\end{document}